\documentclass[journal]{IEEEtran}

\usepackage{graphicx}
\usepackage{url}                 
\usepackage[colorlinks,citecolor=red,urlcolor=blue,bookmarks=false,hypertexnames=true]{hyperref}            
\usepackage{booktabs}            
\usepackage{multirow}            
\usepackage{amsmath,amsfonts,amssymb}
\usepackage{xcolor}

\usepackage{algorithm}
\usepackage{algpseudocode}

\DeclareMathOperator*{\argmin}{arg\,min}

\newcommand{\minikeypoint}[1]{\vspace{0.1cm}\noindent\emph{#1}\quad}

\newcommand{\updated}[1]{\textcolor{black}{#1}}
\newcommand{\cut}[1]{}
\newcommand*\rot{\rotatebox{90}} 

\usepackage{subfiles} 

\begin{document}
\title{Self-Supervised Representation Learning: \\ Introduction, Advances and Challenges}

\author{
    Linus~Ericsson,
    Henry~Gouk,
    Chen~Change~Loy,
    and~Timothy~M.~Hospedales
}

\date{}

\maketitle

\begin{abstract}
    Self-supervised representation learning methods aim to provide powerful deep feature learning without the requirement of large annotated datasets, thus alleviating the annotation bottleneck that is one of the main barriers to practical deployment of deep learning today. These methods have advanced rapidly in recent years, with their efficacy approaching and sometimes surpassing fully supervised pre-training alternatives across a variety of data modalities including image, video, sound, text and graphs.
    This article introduces this vibrant area including key concepts, the four main families of approach and associated state of the art, and how self-supervised methods are applied to diverse modalities of data.
    We further discuss practical considerations including workflows, representation transferability, and compute cost. Finally, we survey the major open challenges in the field that provide fertile ground for future work.
\end{abstract}


\section{Introduction}
\IEEEPARstart{D}{eep} neural networks (DNNs) now underpin state of the art artificial intelligence systems for analysis of diverse data types \cite{He2016DeepRecognition,Vaswani2017AttentionNeed}. However, the conventional paradigm has been to train these systems with supervised learning, where performance has grown roughly logarithmically with annotated dataset size \cite{sun2017unreasonable}. The cost of such annotation has proven to be a scalability bottleneck for the continued advancement of state of the art performance, and a more fundamental barrier for deployment of DNNs in application areas where data and annotations are intrinsically rare, costly, dangerous, or time-consuming to collect.

This situation has motivated a wave of research in self-supervised representation learning (SSRL) \cite{Jing2020Self-supervisedSurvey}, where freely available labels from carefully designed pretext tasks are used as supervision to discriminatively train deep representations. The resulting representations can then be re-used for training a DNN to solve a downstream task of interest using comparatively little task-specific annotated data compared to conventional supervised learning.

Self-supervision refers to learning tasks that ask a DNN to predict one part of the input data---or a label programmatically derivable thereof---given another part of the input. This is in contrast to supervised learning, which asks the DNN to predict a manually provided target output; and generative modelling, which asks a DNN to estimate the density of the input data or learn a generator for input data. Self-supervised algorithms differ primarily in their strategy for defining the derived labels to predict. This choice of \emph{pretext task}, determines the (in)variances of the resulting learned representation; and thus how effective it is for different downstream tasks.

Self-supervised strategies have been leveraged successfully to improve sample efficiency of learning across a variety of modalities from image \cite{Chen2020BigLearners,Grill2020BootstrapLearning,Caron2020UnsupervisedAssignments}, video \cite{Xu2019Self-supervisedPrediction,Alwassel2020Self-SupervisedClustering}, speech \cite{Baevski2020Wav2vecRepresentations,Oord2018RepresentationCoding}, text \cite{Devlin2019BERT:Understanding,Radford2019LanguageLearners} and graphs \cite{Grover2016Node2vec:Networks,Velickovic2019DeepInfomax}. Across these modalities, it can also be applied to boost diverse downstream tasks including not only simple recognition but also detection and localisation \cite{Goyal2019ScalingLearning}, dense prediction (signal transformation) \cite{Goyal2019ScalingLearning}, anomaly detection \cite{georgescu2020anomaly}, and so on. 
\updated{Furthermore, some results suggest that self-supervised representation quality is also a logarithmic function of the amount of unlabelled pre-training data \cite{Goyal2019ScalingLearning}.} If this trend holds, then achievable performance may improve for ``free'' over time as improvements in data collection and compute power allow increasingly large pre-training sets to be used without the need for manually annotating  new data.

\updated{There are various other strategies for improving the data-efficiency of learning, such as transfer learning  \cite{goodfellow2016deepBook,gouk2020distancebased}, semi-supervised learning \cite{englen2020sslSurvey}, active learning and meta-learning. As we shall see, SSRL provides an alternative competitor to conventional transfer learning and semi-supervised learning pipelines; however it can also be complementary to semi-supervised learning and active learning.}

\updated{In this article, we focus on self-supervised algorithms and applications that address learning general-purpose features, or representations, that can be reused to improve learning in downstream tasks.} 
We introduce self-supervised representation learning and review its application and state of the art across several modalities (image, text, speech, graphs, etc), with a specific focus on discriminative SSRL (we exclude generative models such as VAEs, GANs, and Flows; although they can also be used for representation learning). Compared to existing surveys \cite{Jing2020Self-supervisedSurvey}, 
we provide a broader introduction to the field; a wider coverage of different modalities rather than focusing on images; highlight more practical considerations such as representation transferability, compute cost, and deployment strategies; and provide a deeper a discussion of open challenges.

\section{Background}\label{sec:background}

\begin{figure}
    \centering
    \includegraphics[width=0.85\columnwidth]{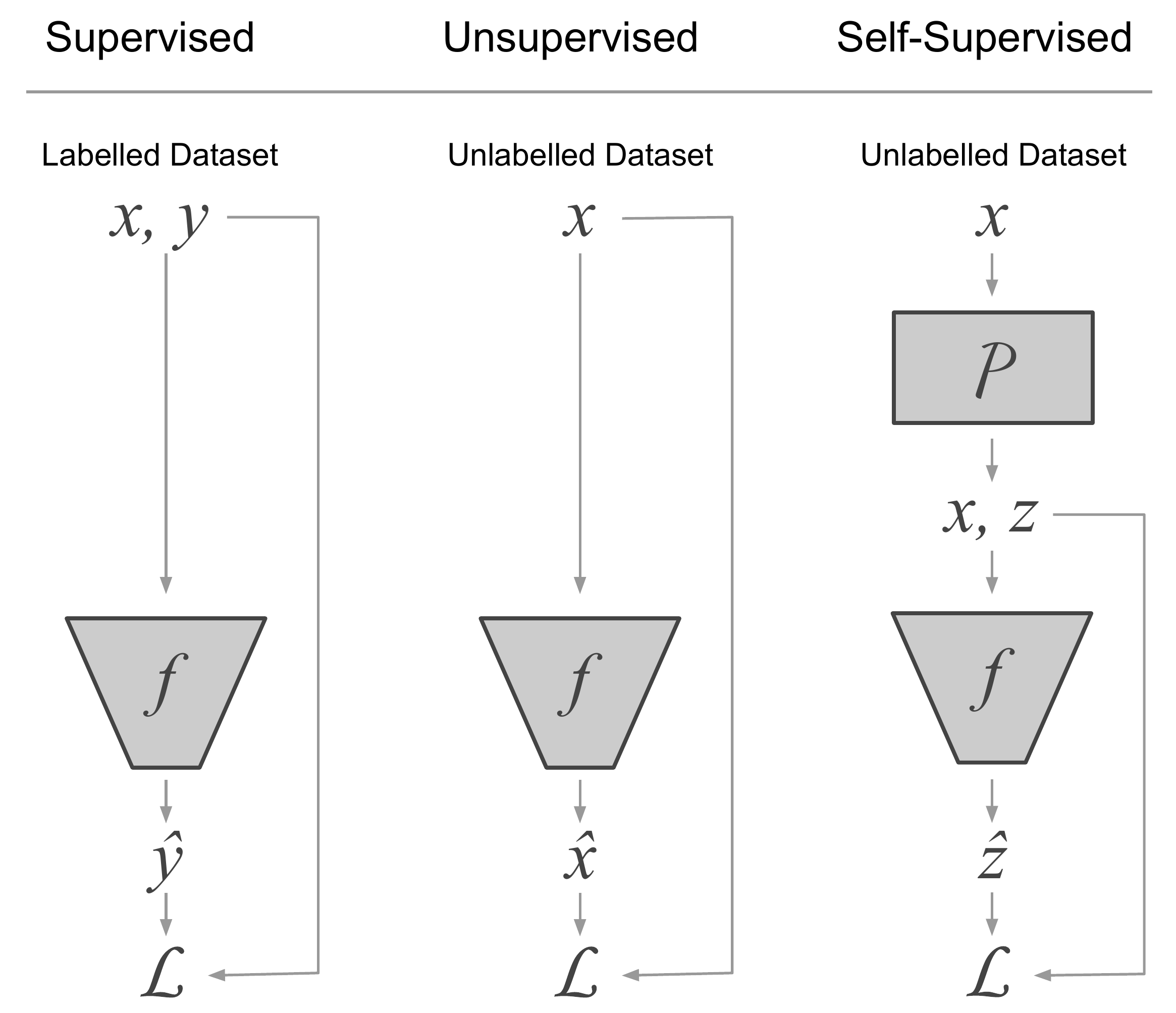}
    \caption{Contrasting supervised, unsupervised and self-supervised learning paradigms for training a model $f$ using raw data $x$, labels $y$, and loss function $\mathcal{L}$. Self-supervision methods introduce pretext tasks $\mathcal{P}$ that generate pseudo-labels $z$ for discriminative training of $f$.}
    \label{fig:paradigms}
\end{figure}

\subsection{Problem Definition}
In this section we introduce the necessary notation for defining the self-supervised representation learning problem and contrast it to other common learning paradigms (Figure~\ref{fig:paradigms}).

\minikeypoint{Supervised Learning} requires a labelled dataset for a target problem we wish to solve $D_t=\{x_i^{(t)},y_i^{(t)}\}^N_{i=1}$, from which we build a predictive model that makes estimates $\hat{y}=f(x)$. In a deep learning context, the predictive model is usually composed of a representation extractor function $h_\theta$ and a classifier/regression function $g_\phi$, $f(x)=g_\phi(h_\theta(x)))$. We train this predictive model by minimising a loss function $\mathcal{L}$ such as the negative log likelihood:
\begin{equation}
    \label{eq:supervised-obj}
    \argmin_{\theta,\phi} \sum_{(x_i^{(t)}, y_i^{(t)})\in D_t} \mathcal{L}({g_\phi(h_\theta(x_i^{(t)}))}, y_i^{(t)}).
\end{equation}
However $h_\theta$ may have hundreds of millions of parameters, requiring millions of labelled data points in $D_t$ to fit this correctly. These millions of annotated data points are not available in most applications, but many do have an essentially free supply of \emph{unlabelled} data points --- as an example consider the wealth of raw audio signal data $x$ vs the limited amount of transcribed speech data $y$ in speech recognition.

\begin{figure*}[t]
    \centering
    \includegraphics[width=1\linewidth]{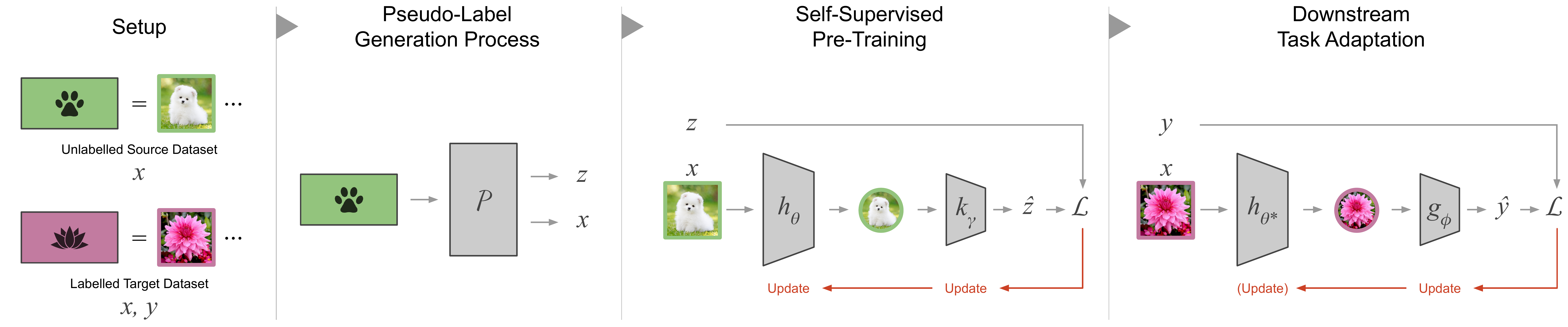}
    \caption{The self-supervised workflow starts with an unlabelled source dataset and a labelled target dataset. As defined by the pretext task, pseudo-labels are programmatically generated from the unlabelled set. The resulting inputs, $x$ and pseudo-labels $z$ are used to pre-train the model $k_\gamma(h_\theta(\cdot))$ -- composed of feature extractor $h_\theta$ and output $k_\gamma$ modules -- to solve the pretext task. After pre-training is complete, the learned weights $\theta^*$ of the feature extractor $h_{\theta^*}$ are transferred, and used together with a new output module $g_\phi$ to solve the downstream target task.}
    \label{fig:ssl_workflow}
\end{figure*}

\minikeypoint{Unsupervised Learning} methods often learn from such unlabelled data by building generative models or density estimators. These range from classic shallow approaches such as Gaussian mixtures \cite{Hastie2009ThePrediction} to deep methods such as variational autoencoders (VAEs) and generative adversarial networks (GANs) \cite{goodfellow2016deepBook}. Other common unsupervised approaches such as autoencoders and clustering \cite{goodfellow2016deepBook} learn compact latent representations. For example autoencoders often optimise a reconstruction objective:
\begin{equation}
    \label{eq:unsupervised-obj}
    \argmin_{\theta,\phi} \sum_{x_i^{(t)}\in D_t} \mathcal{L}({g_\phi(h_\theta(x_i^{(t)}))}, x_i^{(t)}),
\end{equation}
where $h_\theta(\cdot)$ extracts a compact feature from the input, and $g\phi(\cdot)$ uses it to reconstruct the original input.

\updated{
\minikeypoint{Self-Supervised Representation Learning}
 can be seen as a special case of unsupervised learning, since both methods learn without annotations, $y$. While conventional unsupervised methods rely on reconstruction or density estimation objectives, SSRL methods rely on pretext tasks that exploit knowledge about the data modality used for training. }

Although supervised learning methods tend to learn stronger features than unsupervised learning approaches, they require costly and time-consuming work from human annotators to generate the required labels. Self-supervised representation learning techniques aim for the best of both worlds: training a powerful feature extractor using discriminative learning, without the need for manual annotation of training examples.


Given an unlabelled \emph{source} dataset $D_s=\{x_i^{(s)}\}^M_{i=1}$, with $M \gg N$, self-supervised learning addresses how to make use of $D_s$ and $D_t$ together to learn the predictive model $f(x)=g_\phi(h_\theta(x))$.

What defines a self-supervised method is its \emph{pretext} task, consisting of a process, $\mathcal{P}$, to generate pseudo-labels and an objective to guide learning. Given a raw data set like $D_s$, the pretext process programmatically generates pseudo-labels $z$ and possibly modified data points $\{x_i, z_i\}^{M}_{i=1} = \mathcal{P}(D_s)$. As an example, a portion of a speech signal $x$ can be modified by masking out some part of the signal, and the pseudo-label $z$ is defined as the masked out portion of the input. A neural network can then be trained on the objective of predicting the missing portion $z$ given the partially masked $x$.

Most self-supervision research activity addresses deriving pretext tasks $\mathcal{P}$, which enable learning general purpose representations $h_\theta$ that provide high performance and data-efficient learning of downstream tasks $D_t$. Different pretext tasks are discussed in detail in Section \ref{sec:pretext-tasks}. 

The workflow of self-supervision -- also depicted in Fig.~\ref{fig:ssl_workflow}, proceeds as follows:
\begin{enumerate}
    \item The annotated data for the target task forms the dataset $D_t$ and available unlabelled data forms the larger $D_s$.
    \item The pretext task generates a new pseudo-labelled dataset as $\overline{D}_s=\{x_i, z_i\}^{M}_{i=1} = \mathcal{P}(D_s)$ as explained above. \updated{(As the process $\mathcal{P}$ often depends on sampling transformation or masking parameters, it is generally repeated at the start of each epoch of training.)}
    \item The pretext model, $k_\gamma(h_\theta(\cdot))$, is trained to optimise the self-supervised objective on $\overline{D}_s$,
    \begin{equation}
    \label{eq:pretrain}
    \theta^*=\argmin_{\theta,\gamma} \sum_{(x_i,z_i)\in\mathcal{P}(\overline{D}_s)} \mathcal{L}({k_\gamma(h_\theta(x_i))}, z_i). \\
    \end{equation}
    Importantly, this provides a good estimate $\theta^*$ of the potentially hundreds of millions of parameters in $h_\theta$, but without requiring label annotation. 
    \updated{In many cases the input $x_i$ is a single datapoint and the pseudo-label $z_i$ is a class label of scalar value. However, as we will see later in certain types of instance discrimination methods, the input $x_i$ above can consist of multiple datapoints with the pseudo-label $z_i$ describing how the network should relate these datapoints. Similarly, in transformation prediction the input $x_i$ can consist of multiple shuffled chunks while the pseudo-label $z_i$ relates the shuffled order to the original order.}
    \item The pretext output function $k_{\gamma}$ is discarded, and the representation function $h_{\theta^*}$ is transferred as a partial solution to solve the target problem of interest using model $g_\phi(h_{\theta^*}(\cdot))$. Crucially, when representation parameters $\theta^*$ are already well fitted from the self-supervision step in Eq~\ref{eq:pretrain}, only a minority of parameters may need to be learned or refined to solve the target problem, thus enabling it to be solved with a small labelled target dataset $D_t$. There are two common ways to solve the target problem using $\theta^*$ -- fine-tuning and linear readout. 
\end{enumerate}

\updated{The presentation above assumes the target task is labelled and trained with supervised learning, as this is the most common use case. However, unlabelled target tasks like clustering or retrieval can also obviously benefit from self-supervised pre-training if substituted in step 4) above \cite{Mikolov2013DistributedCompositionality}.}
 
\minikeypoint{Linear Readout}
{Let $(\theta, \gamma)$} be the weights of the pre-trained model consisting of a feature extractor $h_\theta$ followed by a task-specific head, $k_\gamma$. The simplest way to re-use $h_\theta$ for a new task is to replace the head with a new one, $g_{\phi}$, designed for the new task. This head is then trained with the feature extractor frozen. Given a target dataset of $N$ instances, $D_{t} = \{x_i^{(t)}, y_i^{(t)}\}_{i=1}^N$, the training objective is
\begin{equation}
    \argmin_{\phi} \frac{1}{N} \sum_{i=1}^N \mathcal{L}(g_{\phi}(h_\theta(x_i^{(t)})), y_i^{(t)}).
\end{equation}
The head is often a simple linear function, leading to the term linear readout. This is often used in the very sparse data regime where the number of unique parameters to learn for the target task must be aggressively limited to avoid over-fitting \cite{Dosovitskiy2014DiscriminativeNetworks,Velickovic2019DeepInfomax,Gao2020GraphTER:Transformations}.

\updated{If enough downstream data is available, it may be better to fit a more complex non-linear function on top of the features. This may consist of multiple linear layers interspersed with non-linearities and potential task-specific modules. In the academic literature, however, it is very common to fit only linear functions in order to simplify comparison between methods.}

\minikeypoint{Fine-Tuning}
Instead of just training a new head, we can retrain the entire network for the new task. We usually still need to replace the pretext head with one suited to the target task, but now we train both the feature extractor and the head,
\begin{equation}
    \argmin_{\theta, \phi} \frac{1}{N} \sum_{i=1}^N \mathcal{L}(g_{\phi}(h_\theta(x_i^{(t)})), y_i^{(t)}).
\end{equation}
Crucially, one must initialise $\theta$ with the values $\theta^*$ obtained during the self-supervised pre-training phase. Given that DNN optimisation is usually non-convex, and assuming a small learning rate, this results in the optimisation above converging towards a local optimum on the target task objective that lies near the local optimum attained for the source task, thus providing knowledge transfer from the pretext source task.

Fine-tuning is often used in the moderately sparse data regime where there is enough target data to at least refine all model parameters; or in regimes where the pretext task/data is not perfectly suited to the downstream task \cite{Chen2020BigLearners,Devlin2019BERT:Understanding,conneau2019xlm,Pathak2016ContextInpainting,Hu2020GPT-GNN:Networks}. \updated{If the source and target domains are well aligned, it may not be necessary -- or even beneficial -- to fine-tune all the parameters. Often only the final few layers of a network need tuning in order to adapt to a new task. In other cases it is enough to tune a specific type of layer, like batch normalisation, to adapt to a slight change in domain.}

In summary, self-supervised representation learning uses unlabelled data to generate pseudo-labels for learning a pretext task. The learned parameters then provide a basis for knowledge transfer to a target task of interest. After pre-training, the transfer can be completed by linear readout or finetuning on the labelled target data.

\subsection{Canonical Use Cases}
When should one consider using self-supervision? SSRL may have diverse benefits in terms of adversarial robustness \cite{hendrycks2019ssl}, model calibration \cite{Ericsson2021HowTransfer}, and interpretability \cite{Ericsson2021HowTransfer} discussed further in Section~\ref{sec:discuss}. However its main use case is to improve data efficiency in situations where there are limited labels for the downstream target task (e.g., semantic segmentation or object detection) and/or domain (e.g., medical or earth observation images) of interest. We mention a few common problem templates and explain how SSRL fits in.
\begin{itemize}
    \item If dense labels are available for the target task and domain, then direct supervised learning may be the most effective approach, and SSRL may not be helpful. 
    \item If the target domain of interest is very different to any available background datasets (e.g., radar vs ImageNet data in imagery), and annotation is expensive in the target domain. Then collecting unlabelled target data for target-domain specific self-supervision, followed by sparse data fine-tuning may be effective. This setting can also be addressed by \emph{semi-supervised} methods \cite{englen2020sslSurvey}, which should then be evaluated as competitors against SSRL.
    \item If the target domain of interest is similar enough to large source datasets (e.g., everyday objects vs ImageNet). Then one can leverage self-supervised pre-training on the source dataset before directly transferring the representation to the target domain of interest. Note that here \emph{conventional supervised} pre-training is a competitor that should be evaluated against SSRL. However, in many cases state of the art SSRL has the edge on supervised pre-training for such transfer settings \cite{Ericsson2021HowTransfer}.
\end{itemize}

\subsection{Deployment Considerations}
In this section we will discuss common ways of using a pre-trained encoder $h_\theta$ for a labelled target dataset. While there are often domain or task-specific methods in the literature for how to best do this, we will focus on some of the most widely adopted approaches.

The target input data is often assumed to lie in the same space as the source data, so that the encoder can be used without modification. The label spaces will most likely differ, however. This means that the head of the pre-trained network, $k_\gamma$ is not suited to solve the target task. The design of the new head $g_{\phi}$ depends mainly on the label space of the target task. For example, in object recognition, the output is likely a vector of class probabilities, for visual object detection additional bounding box locations must be predicted, and for dense prediction a deconvolutional decoder may be introduced.

\begin{figure*}[t]
    \centering
    \includegraphics[width=1.\linewidth]{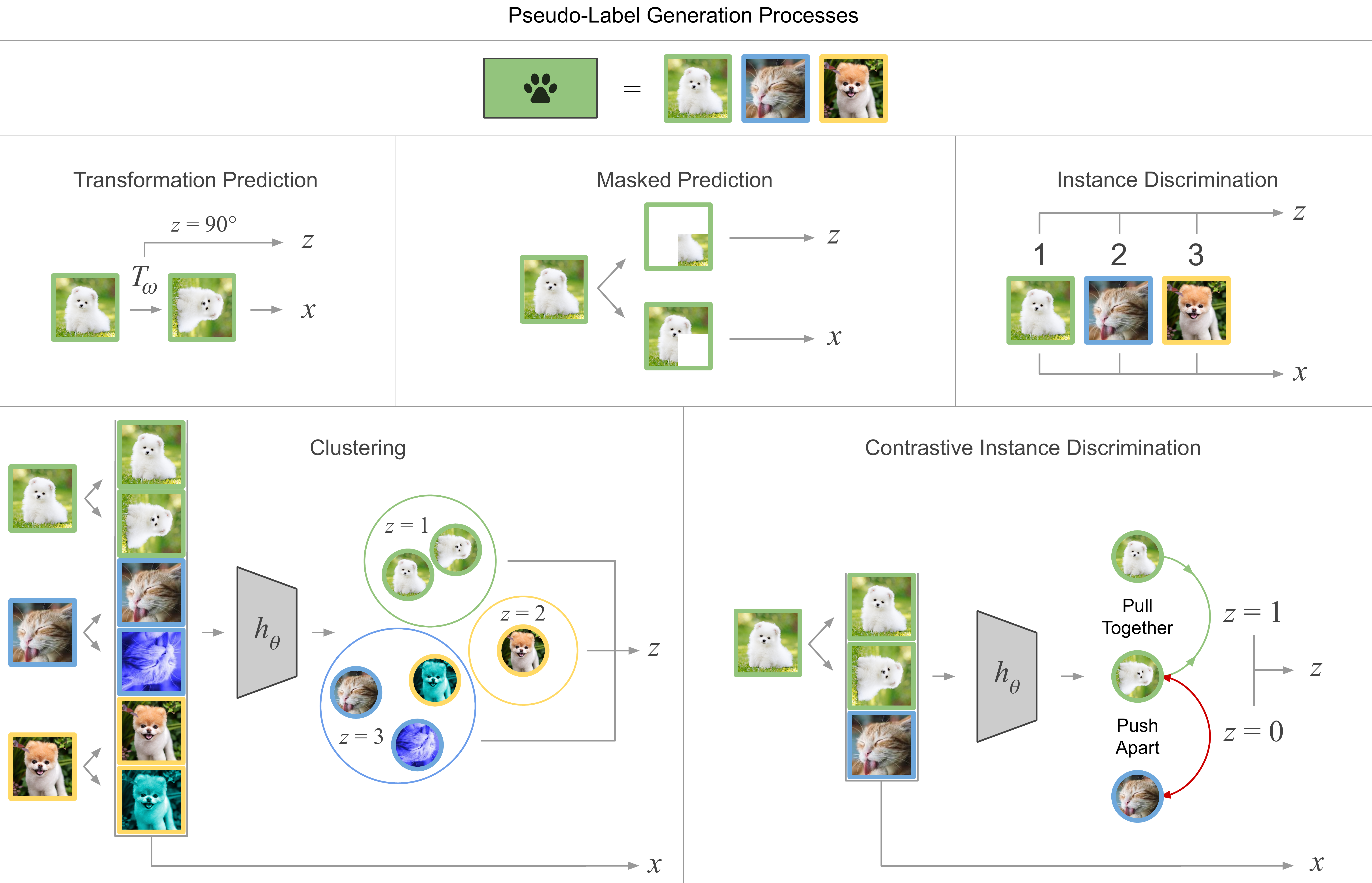}
    \caption{Illustrative examples of the way pseudo-labels are generated in the four families of pretext tasks of our taxonomy: transformation prediction, masked prediction, instance discrimination and clustering. An additional depiction is included of the popular version of instance discrimination using contrastive losses. Squares represent inputs $x$ while circles portray the feature vectors of those inputs, $h_\theta(x)$.}
    \label{fig:pseudo_label_proc}
\end{figure*}

\minikeypoint{Layer Choice}
Given the model pre-trained on the source task, which feature layer is best for extracting features in order to solve a downstream task is an active research question \cite{Goyal2019ScalingLearning}. This is the problem of finding the correct layer to split the encoder $h_\theta$ and the source head $k_\gamma$. The optimal choice can differ from task to task and dataset to dataset, and can involve combining features from several layers, but a general rule is that earlier layers tend to encode simple patterns while later layers can combine these simpler patterns into more complex and abstract representations.

\minikeypoint{Fine-tuning vs Fixed Extractor} An important design choice in deployment phase is whether to fix encoder $h_\theta$, and just train a new classifier module $g_\phi$ using the target data, or fine-tune the encoder while training the classifier. Many SSRL benchmarks use an experimental design relying on linear classifier readout of a frozen encoder. This makes SSRL methods easier to compare due to there being fewer parameters to tune in linear readout. 

There have been mixed results reported in the literature with regards to whether linear readout is sufficient, or whether fine-tuning the entire encoder should improve performance \cite{Kolesnikov2019RevisitingLearning,Ericsson2021HowTransfer}. Which performs better may depend on the amount of available data (fine-tuning is more reliable with more data), the similarity between source and target domain data, and how well suited the (in)variances of the SSRL pretext task used are to the requirements of the downstream task. Conditions with larger domain/task discrepancy are likely to benefit from more fine-tuning.  Of course there are numerous ways to control the amount of fine-tuning allowed in terms of learning rate, and explicitly regularising the fine-tuning step to prevent it over-fitting by limiting the deviation from the initial pre-trained conditions \cite{gouk2020distancebased}.

\minikeypoint{Other Considerations}
A unique issue for SSRL is that it can be difficult to determine the ideal stopping condition for the pretext task as no simple validation signal that can be used. There is not yet an efficient solution for this issue.

Multiple studies have observed that downstream performance after SSRL improves with the capacity of the network architecture used for pre-training \cite{Goyal2019ScalingLearning,Chen2020BigLearners}. While convenient for extracting more performance at the cost of compute and memory, it may create a bottleneck for deploying the resulting fat representation on an embedded or other memory-constrained downstream platform. To alleviate this issue, high-performance and high-parameter count SSRL features can be \emph{distilled} into smaller networks while retaining their good performance \cite{Chen2020BigLearners,Jiao2020TinyBERT:Understanding} (see also Section~\ref{ssec:discussArchitecture}).

\section{Pretext Tasks} \label{sec:pretext-tasks}
In the absence of human-annotated labels, self-supervision uses the intrinsic structure of the raw data and an automated process $\mathcal{P}$ to synthesise a labelled source dataset, $\overline{D}_s = \{x_i, y_i\} = \mathcal{P}(D_s)$. One can then make use of $\overline{D}_s$ as they would any other labelled dataset when pre-training a model, by applying a discriminative supervised learning algorithm. As the pseudo-labels are created from some intrinsic structure in the data, a model learning to predict those labels must recognise and exploit this structure to solve the task successfully. Thus self-supervised algorithm design requires and exploits human prior knowledge about structure in the data to help define meaningful pretext tasks.
Furthermore, different pretext tasks will induce different (in)variance properties in the learned representations, so the choice of method can also be informed by what properties of the representation are required by the downstream task. In this section we divide the various self-supervised pretexts in the literature in four broad families of \emph{masked prediction}, 
\emph{transformation prediction}, \emph{instance discrimination}, and \emph{clustering} -- as illustrated in Figure~\ref{fig:pseudo_label_proc}.

\subsection{Masked Prediction}
\updated{This family of methods is characterised by training the model to \emph{fill in missing data} removed by $\mathcal{P}$. It relies on the assumption that context can be used to infer some types of missing information in the data if the domain is well modelled. Given a raw example, $x_i^{(s)}$, a subset of the elements are extracted to form the pseudo-label, $z^i$, and the remaining components that were not used to create the label are used as the new input example, $x_i$. The pseudo-label generation process therefore looks like $x_i, z_i = \mathcal{P}(x_i^{(s)})$ and is described in full in Alg.~\ref{alg:mp}.}

\updated{
\begin{algorithm}
\caption{Pseudo-label generation process $\mathcal{P}$ for masked prediction}\label{alg:mp}
\begin{algorithmic}
\Require Unlabelled dataset $D_s=\{x_i^{(s)}\}^M_{i=1}$.
\For{$i$ from $1$ to $M$}
\State Generate indices, $I$, of elements to remove from $x_i^{(s)}$
\State $z_i \gets \{ x^{(s)}_{i,j} : j \in I \}$
\State $x_i \gets \{ x^{(s)}_{i,j} : j \notin I \}$
\EndFor
\Ensure $\{x_i, z_i\}^M_{i=1}$.
\end{algorithmic}
\end{algorithm}
}

\updated{As an example of this on real data, a square region of an image can be masked out in the raw example. In this scenario $I$ is the set of indices inside the square mask region, the pixels in the masked region will correspond to $z_i$, and the pixels outside the masked region will be $x_i$. Given $x_i, z_i$, the model can now be trained to minimise e.g., a reconstruction loss like mean squared error,}

\updated{
\begin{equation}
    \label{eq:mp}
    \theta^*=\argmin_{\theta,\gamma} \frac{1}{|\mathcal{P}(D_s)|} \sum_{(x_i,z_i)\in\mathcal{P}(D_s)} \Big({k_\gamma(h_\theta(x_i))} - z_i\Big)^2. \\
\end{equation}
}

A major variant of masked prediction approaches are auto-regressive methods, which treat $x$ as a sequence, and the task is to auto-regressively predict the $t+1$ element of the sequence given the $t$ elements seen so far. By factorizing the joint distribution over $x$ into a product of conditionals, these methods can also be seen as unsupervised generative models.

\minikeypoint{Examples} Common masking methods involve hiding words in sentences for language modelling \cite{Mikolov2013DistributedCompositionality,Devlin2019BERT:Understanding,Brown2020LanguageLearners}, hiding time-slices in speech \cite{Baevski2020Wav2vecRepresentations}, hiding regions of images for inpainting \cite{Pathak2016ContextInpainting}, or hiding edges in graphs \cite{Hu2020GPT-GNN:Networks}. In a multi-modal setting it could correspond to, e.g., predicting the audio signal accompanying a video input or vice-versa. 

\minikeypoint{Considerations} Defining an ideal masking strategy (how much, when and where to mask; which context to provide in predicting the masked information) is important in making effective use of masked prediction. For example, masking too much of a speech signal will make it impossible to infer the missing words, while masking too little of it makes the task too easy to require a rich speech model to be learned.

\subsection{Transformation Prediction}
This family relies on the assumption that inputs have a canonical view and that certain transformations can be applied to that view to change it. The canonical view can for example depend on the effects of gravity in vision (i.e., there is a correct notion of up and down in visual scenes) or temporal ordering in video, speech, or other time-series. Transformation prediction methods apply a transformation that maps from canonical views to alternative views and train the model to predict what transformation has been applied. Given a raw input in its canonical view, $x_i^{(s)}$, a transformation $T_\omega$ is applied to produce $x_i = T_\omega(x_i^{(s)})$, which is fed into the model.

\updated{
\begin{algorithm}
\caption{Pseudo-label generation process $\mathcal{P}$ for transformation prediction}\label{alg:tp}
\begin{algorithmic}
\Require Unlabelled dataset $D_s=\{x_i^{(s)}\}^M_{i=1}$.
\For{$i$ from $1$ to $M$}
\State Sample $\omega \sim \Omega$
\State $x_i \gets T_\omega(x_i^{(s)})$ \Comment{Apply transformation to raw input}
\State $z_i \gets \omega$
\EndFor
\Ensure $\{x_i, z_i\}^M_{i=1}$.
\end{algorithmic}
\end{algorithm}
}

The parameters, $\omega$, of this transform are used as the pseudo-label, $z_i = \omega$, that the model is trained to predict. It is typical for these transformation parameters to be sampled from some distribution, $\Omega$. \updated{The learning objective can be, e.g., a cross-entropy loss in the case of categorical transformation parameters.}

\updated{
\begin{equation}
    \label{eq:mp}
    \theta^*=\argmin_{\theta,\gamma} \sum_{(x_i,z_i)\in\mathcal{P}(D_s)} \mathcal{L}_{CE}\Big({k_\gamma(h_\theta(x_i))}, z_i\Big). \\
\end{equation}
}

\updated{The full process $\mathcal{P}(D_s)$ is described in Alg.~\ref{alg:tp}.} Typically one will generate several different views of each $x_i^{(s)}$, each with a different set of transformation parameters. To succeed, a SSRL method has to learn enough about the latent structure of the data to correctly predict the transformation while being invariant to intra-category variability.

\minikeypoint{Examples} In vision applications, one can apply rotations to the raw images and requiring the network to predict angle of rotation \cite{Gidaris2018UnsupervisedRotations}.
In temporal data, such as videos and other time-series, one can shuffle the temporal order of signal samples, and force the network to predict the original order  \cite{Xu2019Self-supervisedPrediction,Sarkar2020Self-supervisedRecognition}.

\minikeypoint{Considerations} Whatever transformation is chosen, the model will learn to produce representations that are equivariant to that transformation. This is because the information regarding the transformation needs to be retained in the representation in order for the final layer to be able to correctly solve the pretext task. A second consideration is it that depends on data having a canonical view. If there is no canonical view with respect to the set of transformations, then performance will be poor. E.g., satellite or drone earth observation image data may have no canonical view with respect to rotation, so training for rotation prediction on this data may be ineffective.

\begin{figure*}[t]
    \centering
    \includegraphics[width=\linewidth]{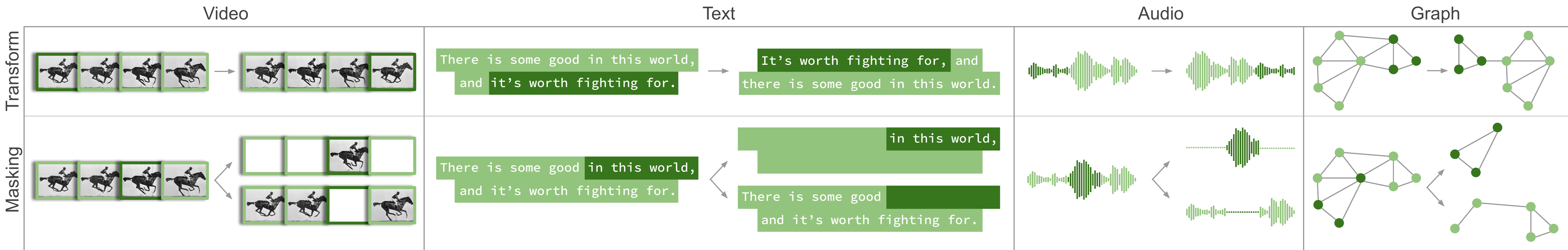}
    \caption{Common transformation and masking methods for different modalities of data. Transformations can be applied to alter the order of sequential data, like the frames in a video, clauses in text or chunks in audio waves. Graphs can be transformed by moving nodes or neighbourhoods. Masking can be applied by hiding frames in videos, groups of words in text, chunks for audio data or subgraphs in graphs. The darker green highlights the portion of the data point that is transformed or masked out. For examples of transforms and masking on image data, see Figure~\ref{fig:pseudo_label_proc}.}
    \label{fig:transforms_and_masks}
\end{figure*}

\subsection{Instance Discrimination}
In this family of methods, each instance in the raw source dataset, $D_s$, is treated as its own class, and the model is trained to discriminate between different instances. There are a few different variations on this framework which we will now describe.

\minikeypoint{Cross-entropy}
\updated{The most straightforward way of tackling instance discrimination is to assign each instance in the dataset a one-hot encoding of its class label, e.g.~instance number 126 in a dataset of 100,000 images would be assigned a vector of length 100,000 with zeros everywhere except for a value of one at position 126. This enables training the network with a categorical cross-entropy loss to predict the correct instances.} This was the approach taken by the early Exemplar-CNN method \cite{Dosovitskiy2014DiscriminativeNetworks}. However, as the size of the dataset grows, the softmax operation to compute class probabilities becomes prohibitively expensive. As such it became difficult to scale this method to large modern datasets where the number of instances---and therefore classes---can be millions \cite{Deng2009ImageNet:Database} or even billions \cite{Mahajan2018ExploringPretraining}. This led to the development of contrastive methods discussed below.

\updated{Another problem within the instance discrimination framework is the lack of intra-class variability. Since each instance in the dataset is treated as its own class, we end up with only a single example of each class. In conventional supervised learning there might be hundreds or thousands of examples within each class to aid the network to learn the inherent variation within in each class. This problem was tackled by Exemplar-CNN via extensive data augmentation. Given a datapoint, we can apply many different transformations to obtain slightly different views of that same datapoint, while preserving its core semantic information. For example, we can slightly change the colour of an image of a car, and it will still be perceived as an image of a car. Figure \ref{fig:transforms_and_masks} shows examples of common transformations across the modalities. The use of data augmentation has become an important component for instance discrimination methods as we will see in the more recent contrastive and regularisation-based methods discussed next.}

\minikeypoint{Contrastive}
\updated{The issue with using a categorical cross-entropy loss to solve instance discrimination is that it becomes intractable for large datasets. Researchers therefore looked for ways to approximate this loss in more efficient ways. The core idea leading recent advances is inspired by metric learning, as well as \cite{hadsell2006} and \cite{Gutmann2010Noise-contrastiveModels}. The idea is to not predict the exact class of the input but to instead predict whether pairs of inputs belong to the same or different classes. This allows the use a binary class label instead of massively high-dimensional class vectors. If a pair of inputs belong to the same class the label is one and if they belong to different classes the label is zero. In this setting, however, the use of data augmentation becomes even more important, as we need to introduce variation between inputs of the same class.}

\updated{
\begin{algorithm}
\caption{Pseudo-label generation process $\mathcal{P}$ for contrastive instance discrimination}\label{alg:id}
\begin{algorithmic}
\Require Unlabelled dataset $D_s=\{x_i^{(s)}\}^M_{i=1}$.
\For{$i$ from $1$ to $M$}
\State Sample $x^a \sim T(x_i^{(s)})$
\State Sample $x^+ \sim T(x_i^{(s)})$
\For{$k$ from $1$ to $K$}
\State Sample $j \sim \mathcal{U}(1, M)$ \Comment{Pick another raw input}.
\State Sample $x^-_k\sim T(x_j^{(s)})$ \Comment{Get a random transform}
\EndFor
\State $x_i \gets \{(x^a, x^+), (x^a, x^-_1), ..., (x^a, x^-_K)\}$.
\State $z_i \gets \{1, 0, ..., 0\}$.
\EndFor
\Ensure $\{x_i, z_i\}^M_{i=1}$.
\end{algorithmic}
\end{algorithm}
}

To formalise the contrastive instance discrimination setup, multiple views of inputs are created via some process $T$ (transformation or sensory-based) and compared in representation space. One input, $x^a \sim T(x^{(s)}_i)$, is chosen to be the \emph{anchor}, and is compared with a positive sample, $x^+ \sim T(x^{(s)}_i)$, which is another view or transform of the same input. The anchor is also contrasted with a negative sample, which is a view of a \emph{different} image, $x^-_j \sim T(x^{(s)}_j)$. \updated{In the context of the general SSRL objective given in Eq.~\ref{eq:pretrain}, this means that the pretext task generator $\mathcal{P}$ produces pretext inputs that each correspond to multiple \emph{pairs} of raw input instances, with associated pseudo-labels indicating whether the pairs are matching or mismatching. See Alg.~\ref{alg:id} for a full description.}

\updated{The samples are then encoded by the feature extractor to obtain their representations, $r^a = h_\theta(x^a)$, $r^+ = h_\theta(x^+)$, $r^-_j = h_\theta(x^-_j)$.} A similarity function $\Phi$ is used to measure the similarity between positive pairs (the anchor with a positive sample) and negative pairs (the anchor with a negative sample). The system is then trained to pull positive pairs closer and push negative pairs apart. A general formulation of the contrastive loss used in many works is
\begin{equation}
    \mathcal{L}_{con} = - \mathbb{E} \left[ \log \frac{\Phi(r^a, r^+)}{\Phi(r^a, r^+) + \sum_{j=1}^k \Phi(r^a, r^-_j)} \right],
\end{equation}
where $k$ different negative samples have been contrasted with the anchor. The model can then be updated by minimising the contrastive loss
\updated{
\begin{equation}
    \label{eq:id}
    \theta^*=\argmin_{\theta,\gamma} \sum_{(x_i,z_i)\in\mathcal{P}(D_s)} \mathcal{L}_{con}\Big({k_\gamma(h_\theta(x_i))}, z_i\Big). \\
\end{equation}
}
Within this framework, methods differ in what similarity function they use, whether they use the same or different encoders for the anchor and other samples, what family of transformations $T$ they use, and how they sample anchor, positive and negative examples. Notable contrastive instance discrimination methods are SimCLR \cite{Chen2020BigLearners} and DGI \cite{Velickovic2019DeepInfomax}.


\minikeypoint{Regularisation-based}
\updated{
While the contrastive framework succeeds in scaling instance discrimination to large datasets, it still has some issues. In order to learn efficiently a very large number of negative examples need to be included in the loss. If we use too few negative examples the network will fail to learn the subtle differences between instances, but too many and training will be computationally expensive. If we were to remove negative examples altogether the features of our network would all collapse to a single constant vector, as there is no incentive to separate features.}

\updated{Regularisation-based approaches to instance discrimination avoid the use of negative examples altogether by regularisation techniques that prevent feature collapse while keeping training efficient. There are many different techniques like using asymmetrical encoding for the two inputs \cite{Grill2020BootstrapLearning} or minimising redundancy via the cross-correlation between features \cite{Zbontar2021BarlowReduction}.}

\minikeypoint{Examples} Established SSRL methods for computer vision including MoCo \cite{He2019MomentumLearning} and SimCLR \cite{Chen2020BigLearners} fall into this family. 
Other applications include speech  \cite{Oord2018RepresentationCoding}; and multi-view \cite{Tian2019ContrastiveCoding} and multi-modal representation learning including audio-visual \cite{Owens2018Audio-VisualFeatures} and visuo-linguistic \cite{radford2021learningCLIP} data
-- where matching and mismatching views of the same instance are contrasted against each other.

\minikeypoint{Considerations} The representations learned here develop high sensitivity to instances, while developing invariance to transformations or views. This means the design of augmentation or view-selection function $T$ is important due to its influence on the invariances learned. For example, aggressive colour augmentation in $T$ may lead to colour invariant representations \cite{Ericsson2021HowTransfer}, which could either be an issue or a benefit depending on the downstream task.
If using different speakers as different views for audio data, the representations would become speaker invariant, which could be beneficial if the downstream task is speech recognition, but an issue if it is speaker diarisation.

Recent work has systematically demonstrated this intuition that the ideal transformations to use do indeed depend on the downstream task \cite{Tian2020WhatLearning}. On one hand this undermines the appealing and widely believed property of SSRL that a single pre-trained model can be re-used for diverse downstream tasks. On the other hand it highlights a new route for research to further improve performance by customising the transformation choice according to the downstream task requirements.

Instance discrimination methods implicitly assume that all instances in the raw dataset represent unique semantic examples, which might not hold -- e.g., if there are many images of the same object. When this assumption is violated, they suffer from \emph{false-positive} pretext task labels \cite{chuang2020debiased}. Nevertheless, they are highly effective in practice despite this violated assumption.

A different issue that is not well understood in theory, but crucial in practice is the sampling and batching strategy for anchor, positive and negative instances for contrastive methods. For example: How to choose negative samples (e.g., at random, via hard negative mining)? What proportion of positive and negative samples, and the size of batches to use \cite{He2019MomentumLearning}? These are all crucial design parameters that vary across the many methods and significantly influence performance.

\subsection{Clustering}
This family of methods focuses on dividing the training data into a number of groups with high intra-group similarity and low inter-group similarity. This relies on the assumption that there exists meaningful similarities by which the data can be grouped -- which is likely the case especially if the data is categorical in nature. There are multiple ways of determining cluster assignment such as connectivity (hierarchical clustering), centroids fitting (e.g.~$k$-means), likelihood maximisation (e.g.~Gaussian mixture modelling) and more \cite{Hastie2009ThePrediction}.

As opposed to traditional clustering, in self-supervised representation learning the aim of the algorithm is to obtain a good feature extractor $f_\theta$ instead of the cluster assignments. Thus one typically jointly performs feature extractor learning and clustering to pre-train the representation prior to downstream use.  This is in contrast to classic clustering methods which usually use a fixed set of features.

A common approach to self-supervised clustering is by alternating two steps, (1) \emph{optimising the clustering objective} by assigning datapoints into clusters based on their representations and (2) \emph{optimising the model} by using the cluster assignments as the pseudo-labels in updates.

A unique feature of the clustering family is thus that the pretext task $\mathcal{P}$ changes during the course of training. Since the pseudo-labels are created by clustering the current representations at each epoch, the labels are updated as the representations change. \updated{This means that the input to the process $\mathcal{P}$ at each iteration is the representations and clusters in addition of the raw data. The full process $\mathcal{P}$ is described in Alg.~\ref{alg:cl}.}

\updated{
\begin{algorithm}
\caption{Pseudo-label generation process $\mathcal{P}$ for clustering}\label{alg:cl}
\begin{algorithmic}
\Require Unlabelled dataset $D_s=\{x_i^{(s)}\}^M_{i=1}$.
\Require Representations $\{r_i\}^M_{i=1}$, where $r_i \gets h_\theta(x_i^{(s)})$
\Require Cluster centres $\{c_j\}^k_{j=1}$, via clustering on $\{r_i\}^M_{i=1}$.
\For{$i$ from $1$ to $M$}
\State Sample $x_i \sim T(x_i^{(s)})$
\State $z_i \gets \argmin_{j \in [k]} \|c_j - r_i\|$
\EndFor
\Ensure $\{x_i, z_i\}^M_{i=1}$.
\end{algorithmic}
\end{algorithm}
}

\updated{Given a cluster assignment, where each input, $x_i$, has its cluster class assigned to $z_i$, we can optimise the model via a cross-entropy loss}
\updated{
\begin{equation}
    \label{eq:mp}
    \theta^*=\argmin_{\theta,\gamma} \sum_{(x_i,z_i)\in\mathcal{P}(D_s)} \mathcal{L}_{CE}\Big({k_\gamma(h_\theta(x_i))}, z_i\Big). \\
\end{equation}
}
\updated{After this, we go back to the clustering step, now using the new representations of our updated model.}

In the cluster assignment step, many works use $k$-means clustering \cite{Caron2018DeepFeatures,Caron2020UnsupervisedAssignments}, where the number of clusters $k$ is a hyperparameter set by evaluating on a validation set of a downstream task. A big problem is that there are degenerate solutions to this, such as assigning all instances to the same cluster \cite{Caron2018DeepFeatures}. To avoid this, methods often enforce that clusters assignments must be balanced  \cite{Caron2020UnsupervisedAssignments}. Recent approaches such as ODC \cite{Zhan2020OnlineLearning} aim to avoid the burden of alternating updates of feature extractor and clusters by simultaneously updating both online.


\minikeypoint{Examples} Major examples include 
DeepCluster, ODC \cite{Caron2018DeepFeatures,Zhan2020OnlineLearning} and SwAV \cite{Caron2020UnsupervisedAssignments} for vision; and XDC \cite{Alwassel2020Self-SupervisedClustering} for multi-modal clustering such as audio and video.

\minikeypoint{Considerations} Many clustering-based SSRL methods \cite{Caron2018DeepFeatures} rely less heavily on data augmentation compared to contrastive methods \cite{Chen2020BigLearners}. This, and avoiding the need to sample triplets, has some benefit in terms of compute cost, but on the other hand the non-stationary nature of the clustering SSRL task (clusters co-evolve with features) imposes additional cost compared to the other pretext tasks with stationary objectives.


Compared to instance discrimination, transformation prediction and masked prediction pretexts, it can be harder to analyse the kinds of (in)variances induced by clustering-based SSRL, making it harder to predict which downstream tasks they are suitable for without empirical evaluation.

\section{Theoretical Underpinning}\label{sec:theory}
The theoretical underpinnings of self-supervised representation learning are lacking compared to standard supervised learning. When analysing a conventional supervised method, the object of most interest is the expected performance of a model on unseen data. The model performance is measured using a task-specific loss function. For example, consider the case of a binary classification problem where the model produces a real-valued score; the sign of this score indicates the predicted class, and the magnitude provides an indication of the confidence with which the model is making the prediction. One loss function that is commonly used for evaluation purposes is the zero--one error,
\begin{equation}
    \mathcal{L}_{0-1}(f, x, y) = \mathbb{I}(f(x)y > 0),
\end{equation}
where $y \in \{-1, 1\}$ is the ground truth label, $f$ is the model, $x$ is an input, and $\mathbb{I}(\cdot)$ is the indicator function. The expected performance of a model on unseen data is then denoted by
\begin{equation}
    \mathbb{E}_{x, y} \lbrack \mathcal{L}_{0-1}(f, x, y) \rbrack.
\end{equation}
The typical goal in statistical learning theory is to bound this quantity from above using the error measured on the training set and some measure of complexity of the class of models, $\mathcal{F}$, that the training algorithm is optimising over. Such bounds are probabilistic, due to the inherent randomness involved in sampling a training dataset, and in some sense can be thought of as sophisticated confidence intervals. These bounds hold uniformly over all $f \in \mathcal{F}$, and usually take the form
\begin{equation}
    \mathbb{E}_{x, y} \lbrack \mathcal{L}_{0-1}(f, x, y) \rbrack \leq \frac{1}{n} \sum_{i=1}^n \mathcal{L}_{0-1}(f, x_i, y_i) + \mathcal{C}(\mathcal{F}, n, \delta),
\end{equation}
where the inequality holds with a probability of at least $1 - \delta$, so $\delta$ is essentially defining the width of a confidence interval as in classic statistical analysis. The complexity term, $\mathcal{C}(\mathcal{F}, n, \delta)$, can be thought of as the upper bound of a confidence interval that takes into account multiple hypothesis testing---i.e., each $f \in \mathcal{F}$ can be thought of as a hypothesis. As more complex classes of models are considered this term will grow larger. Crucially, these bounds assume no knowledge about the underlying data generating distribution, and as such they hold for all distributions.

There are several roadblocks preventing the direct application of this framework to SSRL methods. The most fundamental issue is that the training loss used during self-supervised pre-training measures performance on a pretext task, and is generally not the same loss function used for measuring the performance of the downstream task. As a consequence, the training loss cannot be interpreted as a biased estimate of the expected model performance, and analysis of the model class complexity cannot be used to compensate for the bias in this estimate by widening the confidence interval. A further complication comes from distribution shift. In many cases one wishes to perform self-supervised pre-training on one dataset (such as ImageNet), and then use the resulting features on another dataset with a different marginal distribution. One of the standard assumptions made in learning-theoretic analysis is that elements in the training set and the test set are sampled from the same distribution.

Nevertheless, there is a small but growing literature concerned with theoretical analysis of self-supervised representation learning methods. The key goal these papers share is relating a self-supervised training objective to a supervised objective measured on a small set of labelled data by, e.g., showing that the SSRL loss can be interpreted as an upper bound to a supervised loss. Such analyses typically rely on making assumptions about the data generating process that are hard to verify in practice. We will briefly outline three recent approaches to connecting SSRL with conventional statistical learning theory: one method that applies only to instance discrimination methods~\cite{saunshi2019}, and another that primarily considers how SSRL learns useful representations for natural language tasks~\cite{saunshi2021}\updated{, and finally a paper that makes use of conditional independence to further elucidate how masked prediction pre-text tasks lead to useful representations.}

The analysis of contrastive instance discrimination methods for self-supervised representation learning \cite{saunshi2019} is predicated on the assumption of a specific data generating process. In particular, they assume that the data is generated by a mixture of distributions associated with latent classes. E.g., there is a distribution over the pixels in an image associated with the concept `dog', and there is some prior probability that an image from a particular domain will contain a dog. They demonstrate that one can bound the supervised loss by
\begin{equation}
    \mathbb{E}_{x, y} \lbrack \mathcal{L}_{0-1}(f, x, y) \rbrack \leq \mathbb{E}_{x} \lbrack \mathcal{L}_{ssrl}^-(f, x) \rbrack + s(\rho) + \mathcal{C}(\mathcal{F}, n, \delta),
\end{equation}
where $\mathcal{L}_{ssrl}^-(\cdot, \cdot)$ is a modification to the contrastive loss that considers only negative pairs, and $s(\rho)$ is a function of the mixing coefficients, $\rho$, over the latent classes. This bound relies on $f$ being a centroid classifier on top of the network trained with SSRL, and it is shown that this line of analysis is of limited use on more general families of models.

SSRL on text data is often formalised as a masked prediction problem where, given the first part of a sentence, the task is to predict the next word or remainder of the sentence. Recent work \cite{saunshi2021} has provided a concrete link between the performance on this pretext task and the performance one can expect to see on natural language classification problems. However, their analysis does require an assumption for how classification tasks can be reformulated to make them more comparable with the sentence reconstruction pretext task. Their first contribution is to formalise this assumption as a falsifiable hypothesis and empirically verify that it holds in practice. \updated{Their second main contribution investigates the transfer performance of $\epsilon$-optimal language models---models that achieve an expected cross entropy loss within $\epsilon$ of the expected loss of the best possible model.} They show that, conditioned on this empirically verified hypothesis being true, if one can find a model for next word prediction with an $\epsilon$-optimal cross entropy loss, then the cross entropy loss for a downstream classification task will be $O(\sqrt{\epsilon})$. This implies that developing models that are better at the next word prediction pretext task will translate into better feature representations for natural language classifiers.

\updated{Lee et al. \cite{lee2020predicting} conduct a more general analysis of masked prediction pre-text tasks that is not restricted specifically to the natural language processing domain. Recall that masked prediction pre-text tasks take each source instance, $x_i^{(s)}$, and produce two new objects, $x_i$ and $z_i$, which contain subsets of the elements in the original instance. It is shown in \cite{lee2020predicting} that if there is conditional independence between $x_i$ and $z_i$ given the downstream label (and optionally some additional latent variables), then any model that successfully predicts $z_i$ from $x_s$ must be estimating the label (and optional latent variables). They further generalise their results to the case where one must only assume some notion of approximate conditional independence, which they quantify in terms of covariance matrix norms.}

\updated{While there has been some advances in understanding why contrastive and masked prediction methods can lead to discriminative representations for downstream tasks, this work does rely on assumptions about the data (e.g., conditional independence) that has not been verified to occur in practice. Moreover, the empirical results associated with methods from other parts of our taxonomy, such as transformation prediction and deep clustering, have still not been investigated. An example of how further work could address gaps in our current understanding is to extend theoretical frameworks analysing (shallow) clustering methods \cite{vonluxburg2010clustering} to the deep SSRL paradigm. Future work addressing these limitations would be useful for SSRL researchers, and to the broader AI community that make use of pre-trained features.}

\begin{table}[t]
    \centering
    \caption{Notable methods in each modality. Code/PT: Indicates whether a code-base and pre-trained models are available, respectively. TP: Transformation prediction. MP: Masked Prediction. ID: Instance Discrimination. Cl: Clustering.}
    \label{tab:methods}
    \resizebox{0.8\linewidth}{!}{%
    \begin{tabular}{r|lll}
        {} & Method & Pretext Task & Code/PT \\
        \toprule
        \multirow{8}{*}{\rot{Images}} & RotNet \cite{Gidaris2018UnsupervisedRotations} & TP & \href{https://github.com/gidariss/FeatureLearningRotNet}{Y/Y} \\
         & iGPT \cite{Chen2020GenerativePixels} & MP & \href{https://github.com/openai/image-gpt}{Y/Y} \\
         & Colorization \cite{Larsson2016LearningColorization} & MP & \href{https://github.com/gustavla/self-supervision}{Y/Y} \\
         & Inpainting \cite{Pathak2016ContextInpainting} & MP & \href{https://github.com/pathak22/context-encoder}{Y/Y} \\
         & MoCo \cite{He2019MomentumLearning} & ID & \href{https://github.com/facebookresearch/moco}{Y/Y} \\ 
         & SimCLR \cite{Chen2020BigLearners} & ID & \href{https://github.com/google-research/simclr}{Y/Y} \\
         & BYOL \cite{Grill2020BootstrapLearning} & ID & \href{https://github.com/deepmind/deepmind-research/tree/master/byol}{Y/Y} \\
         & SwAV \cite{Caron2020UnsupervisedAssignments} & Cl & \href{https://github.com/facebookresearch/swav}{Y/Y} \\
         \midrule
        \multirow{4}{*}{\rot{Video/MM}} & VCP \cite{Luo2020VideoLearning} & MP & \href{https://github.com/xudejing/video-clip-order-prediction}{Y}/N \\
         & CLIP \cite{radford2021learningCLIP} & ID & \href{https://github.com/openai/CLIP}{Y/Y} \\
         & XDC \cite{Alwassel2020Self-SupervisedClustering} & Cl & N/\href{https://github.com/HumamAlwassel/XDC}{Y} \\
         & ViLBERT \cite{lu2019vilbert}& MP + ID & \href{https://github.com/facebookresearch/vilbert-multi-task}{Y/Y} \\
         \midrule
        \multirow{4}{*}{\rot{Text}} & word2vec \cite{Mikolov2013DistributedCompositionality} & MP & \href{https://github.com/tmikolov/word2vec}{Y}/\href{https://code.google.com/archive/p/word2vec/}{Y} \\
         & ELMo \cite{Peters2018DeepRepresentations} & MP & \href{https://github.com/allenai/bilm-tf}{Y}/\href{https://allennlp.org/elmo}{Y} \\
         & BERT \cite{Devlin2019BERT:Understanding} & MP & \href{https://github.com/google-research/bert}{Y/Y} \\
         & GPT \cite{Radford2019LanguageLearners,Brown2020LanguageLearners} & MP & \href{https://github.com/openai/gpt-2}{Y/Y} N/N \\
         \midrule
        \multirow{4}{*}{\rot{\quad S \& TS}} & CPC \cite{Oord2018RepresentationCoding} & MP & N/N \\
         & wav2vec          \cite{Baevski2020Wav2vecRepresentations} & MP & \href{https://github.com/pytorch/fairseq/tree/master/examples/wav2vec}{Y/Y} \\
         & STRN \cite{Sarkar2020Self-supervisedRecognition} & TP & \href{https://code.engineering.queensu.ca/17ps21/SSL-ECG}{Y/Y} \\
         \midrule
        \multirow{5}{*}{\rot{Graph}} 
        & Node2Vec \cite{Grover2016Node2vec:Networks} & MP & \href{https://snap.stanford.edu/node2vec/}{Y}/N \\
         & GraphSAGE \cite{Hamilton2017InductiveGraphs} & MP & \href{https://github.com/williamleif/GraphSAGE}{Y}/N \\
         & DGI \cite{Velickovic2019DeepInfomax} & ID & \href{https://github.com/PetarV-/DGI}{Y}/N \\
         & GPT-GNN \cite{Hu2020GPT-GNN:Networks} & MP & \href{https://github.com/acbull/GPT-GNN}{Y/Y} \\
         & GraphTER \cite{Gao2020GraphTER:Transformations} & TP & \href{https://github.com/gyshgx868/graph-ter}{Y/Y} \\
        \bottomrule
    \end{tabular}
    }
\end{table}

\section{Methods and Datasets}
In this section we review major methods and considerations broken down by data modality. Summaries of major methods and datasets for Image, Video, Text, Time-series and Graph modalities are provided in Table~\ref{tab:methods} and \ref{tab:datasets} respectively.

\subsection{Images}
Computer vision tasks performed on still images vary broadly from recognition (whole image classification), detection (object localisaton within an image), and dense prediction (e.g., pixel-wise segmentation). State of the art performance on all of these tasks is achieved by supervised deep learning, and thus SSRL aims to alleviate the annotation bottleneck in computer vision by providing self-supervised pre-training that can be combined with data-efficient fine-tuning.

\updated{
Computer vision has long been dominated by the use of convolutional neural networks (CNNs) that use weight-sharing to reduce the number of learnable parameters by exploiting the spatial properties of images. State of the art architectures usually start with CNN representation encoding $h_\theta(\cdot)$, with ResNet \cite{He2016DeepRecognition} being widely used, before appending task-specific decoding heads $g_\phi$. Many of the initially successful methods in self-supervised representation learning used ResNet backbones \cite{Chen2020BigLearners} but a recent trend has brought Transformer architectures into the vision domain \cite{Chen2020GenerativePixels}. One notable version is the Vision Transformer (ViT) \cite{Dosovitskiy2021AnScale} that is increasingly being used by recent self-supervised methods on image data \cite{radford2021learningCLIP}.}

\subsubsection{Methods} 
All types of pretext tasks (Section~\ref{sec:pretext-tasks}) have been widely applied in still imagery (Table~\ref{tab:methods}). The earliest example of a self-supervised system, given the modern interpretation of the phrase, is the work of \cite{hadsell2006}. This paper introduced two fundamental ideas still relevant to techniques being developed today: (i) metric learning with a contrastive loss and a heuristic for generating training pairs that can be used to train a neural network feature extractor; (ii) using side-information, such as the relative position or viewing angle of training images, can be used to learn invariant or equivariant features. 
Subsequent methods that focused on SSRL for single images also pursued the goal of developing feature extractors that are invariant to different types of transformations, through transformation augmentations~\cite{Dosovitskiy2014DiscriminativeNetworks}.

\begin{table}[t]
    \centering
    \caption{Common source datasets used in each modality.}
    \label{tab:datasets}
    \resizebox{0.9\linewidth}{!}{%
    \begin{tabular}{r|ll|ll}
        {} & Source & Size \\
        \toprule
        \multirow{4}{*}{\rot{Images}} & ImageNet \cite{Deng2009ImageNet:Database} & 1.3M images \\
         & YFC100M \cite{Thomee2015YFCC100M:Research} & 100M images \\
         & iNaturalist \cite{VanHorn2018TheDataset} & 2.7M images \\
         \midrule
        \multirow{4}{*}{\rot{Video/MM}} & Kinetics \cite{Kay2017TheDataset} & 650k videos \\
         & YouTube8M \cite{YouTube-8M:Research} & 8M videos \\
         & HowTo100M \cite{Miech2019HowTo100M:Clips} & 136M videos \\
         \midrule
        \multirow{4}{*}{\rot{Text}} & WikiText \cite{Merity2017PointerModels} & 100M tokens \\
         & OpenWebText \cite{GokaslanOpenWebTextCorpus} & 40GB of text \\
         & Common Crawl \cite{Buck2014N-gramCrawl} & 410B tokens \\
         \midrule
        \multirow{3}{*}{\rot{S/TS}} & LibriSpeech \cite{Panayotov2015Librispeech:Books} & 960 hours of speech \\
         & Libri-light \cite{Kahn2019Libri-Light:Supervision} & 60K hours of speech \\
         & AudioSet \cite{Gemmeke2017AudioEvents} & 5.8k hours of audio \\
         \midrule
        \multirow{4}{*}{\rot{Graph}} & Open Academic Graph \cite{Zhang2019OAG:Graphs} & 178M nodes, 2B edges \\
         & Amazon Review Recommendation \cite{Ni2019JustifyingAspects} & 113M nodes \\
         & PROTEINS \cite{Borgwardt2005ProteinKernels} & 1.1k graphs \\
        \bottomrule
    \end{tabular}
    }
\end{table}

Several methods fall into the transformation prediction family, focusing on modifying unlabelled images using a known transformation, like rotation \cite{Gidaris2018UnsupervisedRotations}, and then training the network to predict the angle of that rotation. 
Others mask out information in the training images and require the network to reconstruct it, leading to pretext tasks such as colourisation \cite{Larsson2016LearningColorization} and inpainting \cite{Pathak2016ContextInpainting}, where colour channels and image patches are removed, respectively. A state of the art example in this category is iGPT \cite{Chen2020GenerativePixels}, which exploits a self-attention architecture and masked prediction for representation learning. 

\begin{figure}[t]
    \centering
    \includegraphics[width=1.0\linewidth]{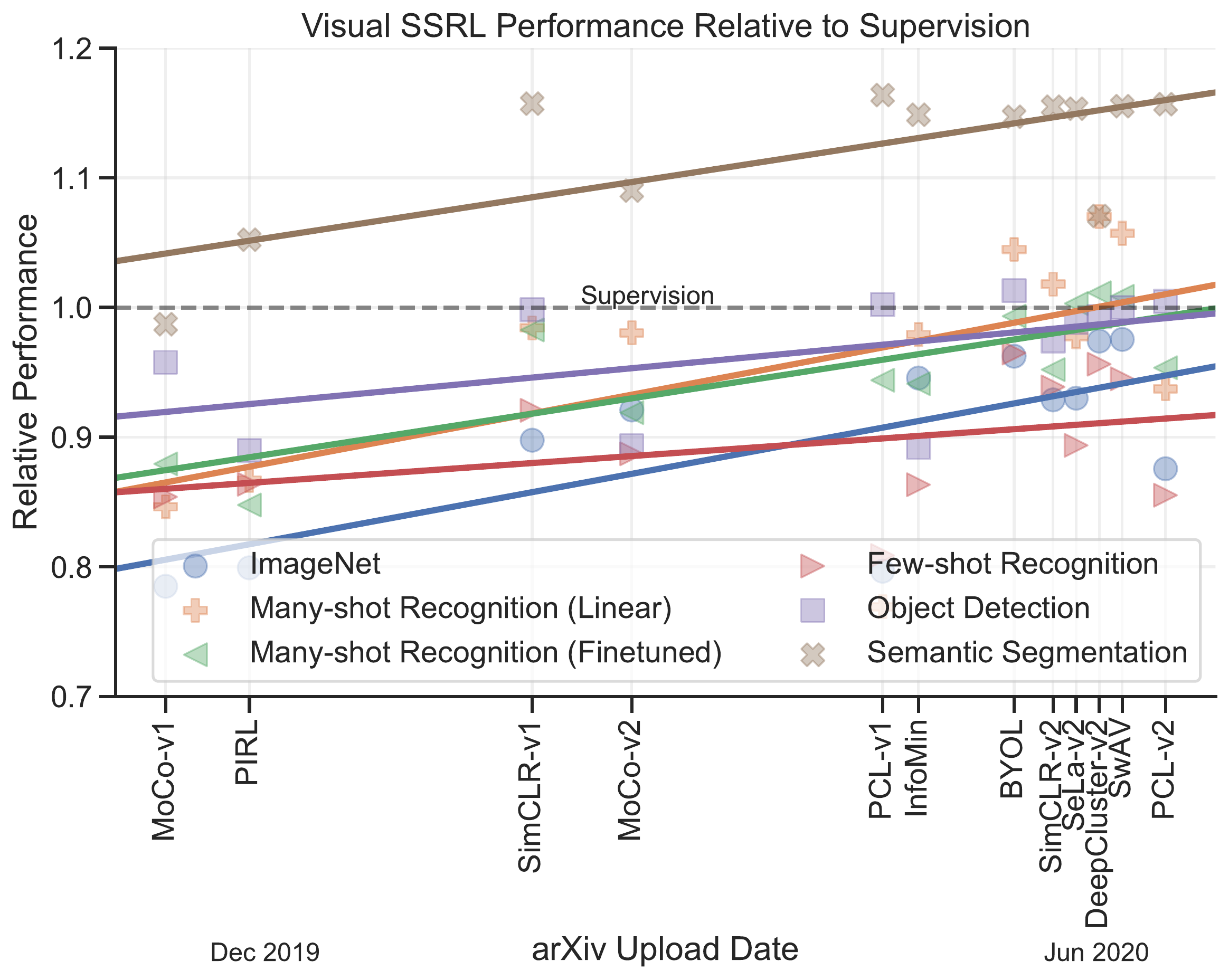}
    \caption{The relative performance of SSRL methods on visual tasks, compared to a supervised baseline. Figure produced based on results in \cite{Ericsson2021HowTransfer}.}
    \label{fig:visual_performance}
\end{figure}

The majority of recent methods focus on the relations between different images in the dataset, using instance discrimination \cite{Chen2020BigLearners,He2019MomentumLearning} or clustering \cite{Caron2020UnsupervisedAssignments}, and heavy data augmentation has become a vital component required by all methods to achieve high performance. Progress has accelerated rapidly in the last two years, with the latest methods now systematically outperforming supervised pre-training in diverse downstream tasks and datasets \cite{Ericsson2021HowTransfer}, as shown in Figure \ref{fig:visual_performance}.\cut{ Among current state of the art clustering and instance discrimination methods are best, with DeepCluster-v2 coming out top in recognition, and SimCLR-v2 leading in detection and dense prediction \cite{Ericsson2021HowTransfer}.}

\begin{figure}[t]
    \centering
    \includegraphics[width=1.0\linewidth]{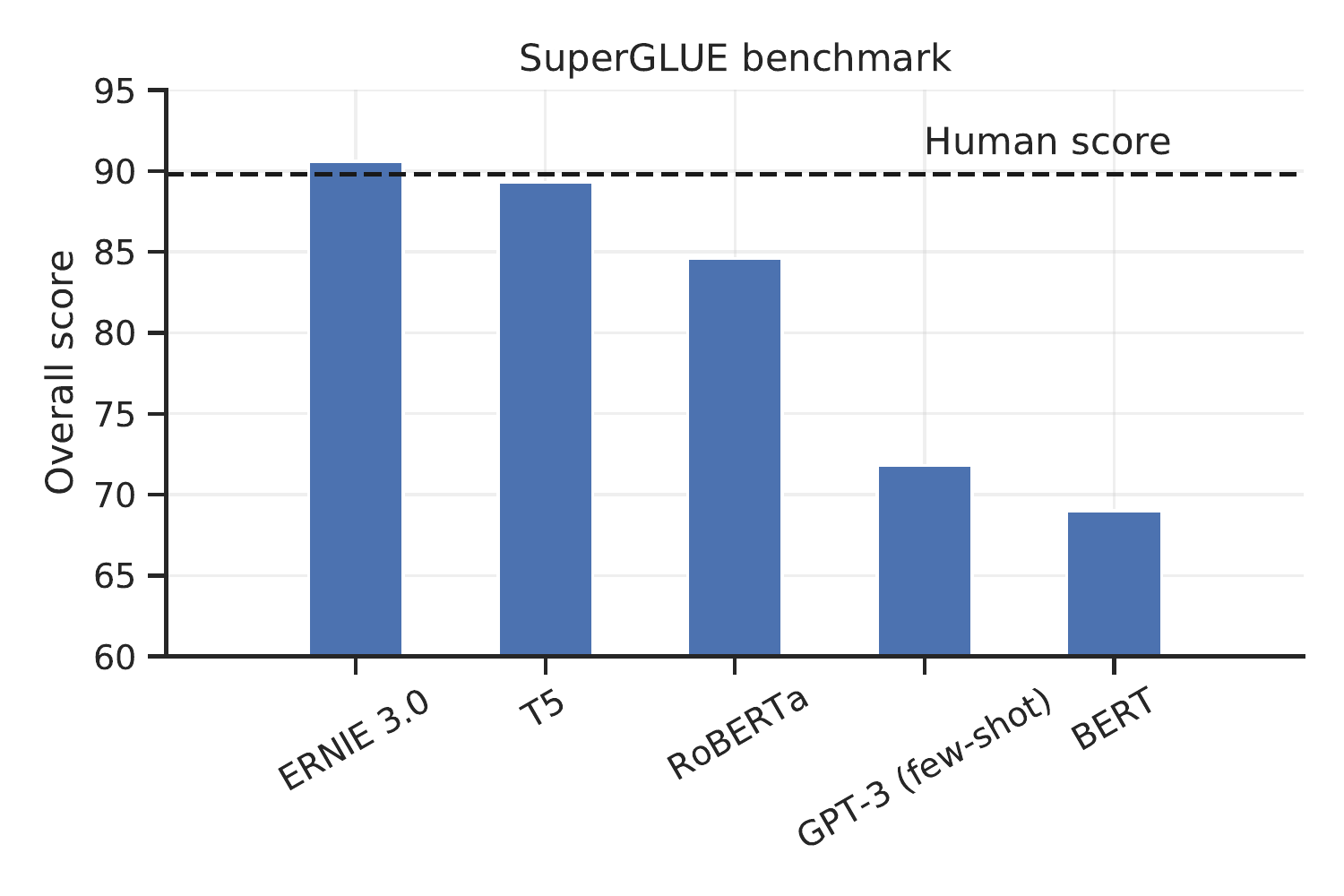}
    \caption{The performance of SSRL methods on the textual benchmark SuperGLUE, compared to a baseline of human performance. Selected methods taken from the official leaderboard at \url{https://super.gluebenchmark.com/leaderboard}.}
    \label{fig:text_performance}
\end{figure}

\subsubsection{Datasets}
As in much of computer vision, ImageNet \cite{Deng2009ImageNet:Database} is the most common source dataset for self-supervised pre-training \cite{He2019MomentumLearning,Chen2020BigLearners,Caron2020UnsupervisedAssignments,Grill2020BootstrapLearning}, consisting of 1.28 million training images across 1,000 object categories, with the most commonly used resolution at $224 \times 224$. 
Many methods are increasingly using datasets much larger than ImageNet. For example YFCC100M \cite{Thomee2015YFCC100M:Research}, with 100 million images from Flickr, used by \cite{Caron2018DeepFeatures}, and \cite{Mahajan2018ExploringPretraining} with 3.5 billion images from Instagram. Subsets of the latter are used by \cite{He2019MomentumLearning,Caron2020UnsupervisedAssignments}.


The ImageNet benchmark is a highly curated dataset, with certain biases that do not appear in natural images, such as centring of objects and clear isolation of object from background. 
iNaturalist \cite{VanHorn2018TheDataset} is a collection of wildlife datasets compiled by a citizen science project, where members upload their own photographs and others collectively annotate them. This forms a more natural dataset which exhibits class imbalance and distractor objects which often complicate real-world tasks. While it has not yet served as the source dataset for any new method, it has been used to benchmark the robustness of existing SSRL methods to more uncurated data  \cite{MacAodhaBenchmarkingCollections}.

\cut{
\minikeypoint{Target}
There are diverse downstream tasks of interest in the image modality, which usually require different target datasets. The most common downstream task is object recognition. Here we find datasets covering coarse-grained (CIFAR10/CIFAR100 \cite{Krizhevsky2009LearningImages}, Caltech-101 \cite{Fei-Fei2004LearningCategories}) and fine-grained classification (iNat \cite{VanHorn2018TheDataset}, FGVC Aircraft \cite{Maji2013Fine-GrainedAircraft}\cut{, Food-101 \cite{Bossard2014Food-101Forests}v, Oxford-IIT Pets \cite{Parkhi2012CatsDogs}, Oxford-VGG Flowers}), and scenes (Places205 \cite{Zhou2017Places:Recognition}, SUN397 \cite{Xiao2010SUNZoo}).

For the object detection task, two datasets, the smaller PASCAL VOC \cite{Everingham2010TheChallenge} and the larger Microsoft COCO \cite{Lin2014MicrosoftContext} are by far the most popular benchmarks.

For predicting per-pixel characteristics at an object-level, there are the instance segmentation datasets Microsoft COCO \cite{Lin2014MicrosoftContext}, Cityscapes \cite{Cordts2016TheUnderstanding} and NYU-Depth-v2 \cite{Silberman2012IndoorImages}. For semantic segmentation there are PASCAL VOC \cite{Everingham2010TheChallenge} and Cityscapes \cite{Cordts2016TheUnderstanding}. Other common per-pixel prediction tasks are depth and surface normal estimation with the most popular dataset being NYU-Depth-v2 \cite{Silberman2012IndoorImages}.
}

\subsubsection{Applications}
On established benchmarks, SSRL has had widespread and significant success in matching and surpassing supervised pre-training performance, especially for image recognition tasks, and in photo imagery of similar character to ImageNet (Figure~\ref{fig:visual_performance}). Progress in transfer to more diverse downstream tasks such as detection and segmentation; as well as downstream datasets which are out of distribution with respect to pre-training data has also been steady \cite{Ericsson2021HowTransfer}, if less rapid.

Beyond common benchmarks, SSRL has been successfully applied in application areas where labelled data is sparse, such as earth observation remote sensing \cite{tao2020remote}. 
In these cases, pre-training on the available unlabelled target-domain data was beneficial to compensate for sparse annotations. A growing downstream consumer of SSRL is the medical imaging domain, where labelled data is often intrinsically sparse or too expensive to collect in bulk for end-to-end learning from scratch. For example \cite{chen2019self} used unlabelled brain scan images to perform image restoration (an inpainting-like task), improving upon random initialisation for fine-tuning several downstream tasks. A somewhat unique feature of the medical imaging domain is the processing of 3D volumetric images such as from MRI images. This has also recently inspired various extensions of standard pretext tasks into 3D \cite{taleb20203d}.

A final application where SSRL pre-training has been successfully applied is that of anomaly detection. SSRL-based approaches typically either train a feature to be used in conjunction with a classic generative anomaly detector, or more interestingly use the SSRL objective itself to produce an anomaly detection score. For example, current state of the art anomaly detectors \cite{golan2018deepAnomaly} rely on SSRL training of rotation prediction, with the rotation prediction accuracy providing the anomaly score. \cut{To use this model to evaluate putative anomalies, images are rotated and then evaluated by the rotation prediction network, with the rotation prediction accuracy providing the anomaly score \cite{golan2018deepAnomaly}. Such an anomaly score is understandable, as intuitively the network will be better at predicting rotation for images similar to those seen during training, than for novel images types.}

\subsection{Video and Multi-modal}
In the domain of video and multi-modality diverse tasks are of interest including video recognition, action/event detection (localisation of an event within a longer video), tracking (localising an object within frames and across time), and cross-modal retrieval (e.g., retrieving a video frame given associated subtitles). State of the art architectures again dominate all of these tasks given access to sufficient training data to train encoder and task-specific decoder components.

Common architectures  $h_\theta(\cdot)$ for encoding videos include 3D CNNs or multi-stream encoders that process appearance and motion separately. In the case of multi-modal processing of video and audio, or video and associated text, one requires a synchronised video CNN encoder as well as a text/audio encoder (e.g., recurrent neural network) to encode the multi-modal streams. These data streams may then be fused into a single representation and decoded at each time-step (e.g., for localisation/detection), or first pooled over time (e.g., for video-level recognition).

\subsubsection{Methods} 
Transformation prediction (TP) and contrastive instance discrimination methods are the most widely used for SSRL in video. 
There are a wide variety of TP pretexts in video. Rotation, and colorization discussed earlier are also widely generalised to video data. Making more unique use of the temporal nature of video, one can for example predict the ordering of frames or clips \cite{Xu2019Self-supervisedPrediction},  or the speeding up or slowing down of videos.

In terms of contrastive instance discrimination methods, data augmentation has been the main mode of obtaining different views in still-imagery. However, for videos several methods exploit multiple sensory views, like RGB, optical flow, depth, and surface normals \cite{Tian2019ContrastiveCoding,Han2020Self-supervisedLearning} that provide different views for learning cross-view video clip matching.

A recent notable method in the instance discrimination family is CLIP \cite{radford2021learningCLIP}, a visuo-linguistic multi-modal learning algorithm that has further advanced state of the art in robust visual representation learning by crawling pairs of images and associated text from the internet, and exploiting them for cross-view contrastive learning. Massive multi-modal pre-training was shown to lead to excellent performance on diverse downstream tasks including language-based image retrieval.

Clustering has been used in similar ways to match inputs from different modalities to the same clusters \cite{Alwassel2020Self-SupervisedClustering}. Finally, masked prediction has been applied through
filling in masked out clips \cite{Luo2020VideoLearning}.

\subsubsection{Datasets}
There are several datasets of videos used for pre-training in this modality. Kinetics \cite{Kay2017TheDataset} is a large action recognition dataset of human-object and human-human interactions, collected from YouTube videos. One version, Kinetics-400, contains around 300k videos. There are larger versions of the dataset with up to 700 classes and 650k videos. Recently, a group of very large-scale datasets have been constructed from publically available videos on social platforms, like YouTube8M \cite{YouTube-8M:Research} 
and HowTo100M \cite{Miech2019HowTo100M:Clips}, the latter containing 136 million YouTube instructional videos with narration with captions across 23k visual tasks. 

For methods using multiple modalities, the visual and audio information often come from the large video datasets discussed above \cite{Alayrac2020Self-SupervisedNetworksb}. An additional dataset that has been considered here is AudioSet \cite{Gemmeke2017AudioEvents}, an audio event detection dataset. For methods using text information, this often obtained from automated transcription using ASR. Other datasets have textual information built in, such as subtitles or video descriptions.

\cut{
\minikeypoint{Targets}
The main tasks considered in this modality are action recognition and video retrieval on two datasets. UCF101 \cite{Soomro2013UCF101:Wild} is an action recognition dataset consisting of 13k videos over 101 action classes. The second is HMDB51 \cite{kuehne2011hmdb} with 7k videos over 51 action classes. Other tasks explored in the literature are captioning on YouCook2 \cite{zhou2018youcook2}, segmentation on COIN \cite{tang2019coin}, unintentional action recognition on Oops! \cite{Epstein2020OopsVideo}, and next action prediction on ActivityNet \cite{caba2015activitynet}\cut{, Breakfast \cite{Kuehne2014TheActivities} and 50Salads \cite{Stein2013CombiningActivities}}.

Multi-modal methods tend to evaluate on the same action recognition benchmarks as discussed above. However, they have additionally considered sound classification on datasets like ESC-50 \cite{piczak2015esc}\cut{ and DCASE \cite{Stowell2015DetectionEvents}}.
}

\subsubsection{Applications}
As outlined in the previous section, the most common application and benchmark scenario for video SSRL is in video action recognition and detection in various guises. SSRL has made rapid progress in this area and state of the art methods trained on massive pre-training sources lead to significantly better performance than direct training on an array of standard benchmarks \cite{Alwassel2020Self-SupervisedClustering}, but do not yet reliably surpass supervised pre-training on the same source datasets as in the case of still images earlier.

Similarly to the still image domain SSRL has been successfully applied to video anomaly detection. For example, given a TP pretext task of arrow of time prediction (differentiating forward vs reverse frame sequences) among others, videos with high probability of being reversed can be considered anomalous \cite{georgescu2020anomaly}.

\cut{
Self-supervised representation learning has been exploited as pre-training prior to several variants of tracking in videos. One family of methods learns  \cite{Vondrick2018TrackingVideos,wang2019cycleTime} a pixel-wise representation suitable for computing dense correspondence across-frames based on which keypoints, human skeletons, instance segmentations, and object segmentations can be tracked in time \emph{without} any task-specific fine-tuning. This can be achieved for example by using cross-frame colorization as a pretext task \cite{Vondrick2018TrackingVideos}. Of course this is extremely valuable as supervised annotation of frame-wise segmentation etc for conventional supervised learning is incredibly expensive. More conventionally with respect to the pipeline outlined in Section~\ref{sec:background}, one can use SSRL such as contrastive instance discrimination to learn a representation \cite{gordon2020watching}, upon which a conventional siamese object tracker can be trained. 
}

Video data is often multi-modal, covering RGB+D, video+audio, or video+text (e.g., from script or text2speech) modalities. 
It is noteworthy that several studies \cite{Tian2019ContrastiveCoding,Alayrac2020Self-SupervisedNetworksb} have explored how SSRL on multi-modal source data can be used to learn a stronger representation for single-modality downstream tasks, and ultimately outperform single-modality pre-training on diverse downstream tasks in uni-modal video, still-image, or audio domains \cite{Alayrac2020Self-SupervisedNetworksb}.

With regards to the video and text, several recent SSRL studies have learned joint multi-modal representations. 
Notably, ViLBERT \cite{lu2019vilbert}, exploited both BERT-like masked prediction and contrastive instance discrimination to learn a multi-modal representation which then achieved state of the art performance in downstream vision and language tasks such as caption-based retrieval, visual question answering, and visual commonsense reasoning.

\subsection{Text and Natural Language}
Natural Language Processing (NLP) methods aim to learn from raw input text and solve a wide variety of tasks ranging from low-level such as word-similarity, part of speech tagging and sentiment; to high-level tasks such as question answering and language translation. State of the art approaches are often based on deep sequence encoders such as LSTM, and in recent years self-attention based approaches have dominated \cite{Vaswani2017AttentionNeed}. With data annotation being a major bottleneck, NLP was the first disciplines to make major and successful use of self-supervision \cite{Mikolov2013DistributedCompositionality}. 

\subsubsection{Methods}
Self-supervised representation learning has been a fundamental component in natural language processing (NLP) for many years. Masked prediction methods have been particularly effective in this modality, with word embeddings becoming widely adopted as they succeed in producing representations that capture the semantic similarity of words, as well as being able to deal with arbitrary vocabulary sizes. word2vec \cite{Mikolov2013DistributedCompositionality} and related methods 
work by either predicting a central word given its neighbours -- called \emph{continuous bag of words} (CBOW) -- or predicting the neighbours given the central word -- called \emph{skip-gram}. Given such pre-trained word embeddings, a target task is then solved by mapping input tokens to their vector embeddings and learning a model on top of them. Since the embedding for a word is fixed after training, it cannot adapt to the context in which the word appears, causing a problem for words with many meanings.

As opposed to these non-contextual embedding methods, topical contextual methods learn embeddings which change depending on the surrounding words. The two most common approaches to this are next word prediction \cite{Radford2019LanguageLearners} and masked word prediction \cite{Devlin2019BERT:Understanding}, with the landmark BERT method combining the latter with next sentence prediction \cite{Devlin2019BERT:Understanding}. For the encoder architecture, recurrent networks like LSTMs \cite{goodfellow2016deepBook} have long been used to model the context while recent works have moved to Transformer-based architectures with self-attention \cite{Vaswani2017AttentionNeed}, which allow longer range connections to be made across words in a sentence, but require more data for training. 
A final trend is that new models are becoming bigger and bigger, counting ELMo \cite{Peters2018DeepRepresentations} at 94M, BERT \cite{Devlin2019BERT:Understanding} at 340M, GPT-2 \cite{Radford2019LanguageLearners} at 1.5B and GPT-3 \cite{Brown2020LanguageLearners} at 175B parameters. \updated{Recent progress on this type of large-scale masked prediction has led to performance surpassing human baselines on language understanding tasks. This can be seen in Figure \ref{fig:text_performance} where we show the performance of selected top models from the leaderboard of the common SuperGLUE \cite{Wang2019SuperGLUE:Systems} benchmark.}

\cut{
The community has acknowledged this scalability limitation, and has thus motivated a newer line of work that strives to improve the parameter efficiency of these models while maintaining high performance, like 
ConvBERT \cite{Jiang2020ConvBERT:Convolution} and TinyBERT \cite{Jiao2020TinyBERT:Understanding}. For example the recent ConvBERT \cite{Jiang2020ConvBERT:Convolution} reduces 50-fold the FLOPs required to train BERT \cite{Devlin2019BERT:Understanding} from $6.4 \times 10^{19}$ to $1.3 \times 10^{18}$ while maintaining similar performance. Nevertheless, work remains to be done as even this largely reduced cost is prohibitive for most users.
}

All methods discussed above belong to the masked prediction family of methods and they have been the most successful and widely adopted. But there are examples of transformation prediction such as 
recovering the order of permuted \cite{Lewis2020BART:Comprehension} or rotated \cite{Lewis2020BART:Comprehension} sentences. These have often been used as complementary signals in order to improve downstream performance on a particular task. 

\subsubsection{Datasets}
Self-supervision in language has shown to benefit from ever larger corpora of text. This has led to huge datasets being created, primarily by crawling the web for the data. Early word embeddings made heavy use of 
Wikipedia articles \cite{Merity2017PointerModels},  
or crawls of news sites and social media sites like Twitter. As models have become larger and require more text to train on, the organisations training these models have begun using private datasets which are not publically available \cite{Radford2019LanguageLearners,Brown2020LanguageLearners}. Attempts have been made at replicating the data used in such papers, for example the OpenWebText Corpus \cite{GokaslanOpenWebTextCorpus}. Another example is Common Crawl \cite{Buck2014N-gramCrawl}, a non-profit project that makes data from billions of web pages freely accessible. Various datasets have been created from this data 
and filtered versions of the entire corpus often form the bulk of training sets \cite{Brown2020LanguageLearners}. Using a combination of the above data sources, the total size of the training set used in state-of-the-art language modeling is now on the scale of 500 billion tokens \cite{Brown2020LanguageLearners}.

\cut{
\minikeypoint{Target}
As self-supervision has been a foundation of NLP for several years now, the field has matured and somewhat converged to a set of standardised benchmarks covering a broad range of downstream tasks and text domains. The Stanford Question Answering Dataset benchmark (SQuAD) \cite{Rajpurkar2016SQuAD:Text},
tests reading comprehension with questions on Wikipedia articles. ReAding Comprehension from English Examinations (RACE) \cite{Lai2017RACE:Examinations} also tests comprehension on questions from English exams originally created for middle and high school students. General Language Understanding Evaluation (GLUE) \cite{Wang2019GLUE:Understanding} is a suite of nine sentence or sentence-pair tasks, covering a diverse range of text domains, dataset sizes, and difficulties. It has recently been replaced by a newer incarnation of the benchmark, SuperGLUE \cite{Wang2019SuperGLUE:Systems}.
}

\subsubsection{Applications}\label{ssec:text:apps}
{SSRL has made a major impact on a host of problems involving multiple languages, which introduces a new kind of source/target dichotomy besides the task and domain-level dichotomies we have focused on thus far.} In the simplest (within-language) scenario, SSRL can benefit all the standard language understanding tasks (classification, QA, etc) for \emph{low resource languages}. One can pre-train SSRL models on a large corpora of high-resource languages before fine-tuning them on smaller corpora of low-resource languages \cite{conneau2019xlm}. For cross-language tasks such as machine translation, one can pre-train SSRL models (e.g., for masked prediction) that are multi-lingual, in that they simultaneously encode/decode data from more than one language. These multi-lingual language models are then well primed for comparatively low-data fine-tuning for translation \cite{conneau2019xlm},
or provide good representations to drive unsupervised \cite{conneau2019xlm} learning of translation models. This is valuable, as vanilla translation models are extremely expensive to supervise due to requiring a vast number of aligned (translated) sentence pairs across languages. 

The conventional task-specific fine-tuning as outlined in Section~\ref{sec:background} is the dominant paradigm for exploiting SSRL in language. However, A notable exception to this is in the recent GPT-3 \cite{Brown2020LanguageLearners} language model. A key observation in this work is that a sufficiently scaled-up 175B parameter generative language model can often perform few or zero-shot learning of a new task in a purely feed-forward manner (no back-propagation or fine-tuning) simply by prompting the model with the few-training examples, and the query and allowing it to complete the answer. 

\subsubsection{Considerations} A growing concern in language modeling is the extent to which biases implicit in the large training corpora for SSRL become baked into the resulting language models, for example sexist or racist stereotypes. Vast corpora must be used for SSRL, so training data cannot be filtered for appropriateness manually. A small but growing body of work aims to develop SSRL variants with reduced bias \cite{huang2019reducing}.

\subsection{Audio \& Time-series}
Classic approaches to audio analysis tasks such as speech recognition compute  mel-frequency cepstrum coefficients (MFCC) from the raw audio data and then models the sequence via Gaussian Mixture Models and Hidden Markov Models. Meanwhile contemporary neural network approaches trained by supervised learning have dominated in settings where massive annotated training data is available \cite{amodei2016deepSpeech2}. Against this backdrop, self-supervised methods have very recently made massive advances in alleviating this annotation bottleneck, enabling state of the art audio analysis methods to be trained with relatively sparse annotations. 

Self-supervised methods in the audio analysis arena have exploited architectures $h_\theta$ spanning all the popular options for time-series data including recurrent  \cite{Chung2019AnLearning}, convolutional \cite{Oord2018RepresentationCoding} and self-attention \cite{Baevski2020Wav2vecRepresentations} networks. These are usually applied directly to raw waveform data to build a representation without any pre-processing step such as MFCC.

\subsubsection{Methods}
In terms of self supervision algorithms, numerous studies have successfully adapted the insights of self-attention based language models \cite{Devlin2019BERT:Understanding} to audio data. As a pretext task, random segments of the input sequence are masked and predicted by a self-attention architecture. However, a key difference is that language models work with discrete token sequences -- thus enabling the pretext to be formalised as multi-class classification, while audio time-series are naturally continuous. Thus solutions to formalising a masked prediction task for audio have either quantised the speech embedding for classification -- such as Wav2Vec-2.0 \cite{Baevski2020Wav2vecRepresentations}; applied contrastive losses to differentiate the masked segment from alternative distractors -- such as CPC  \cite{Oord2018RepresentationCoding}; 
or replaced classification-based prediction with regression layers to directly synthesise the masked frame -- such as APC \cite{Chung2019AnLearning}. Other approaches such as PASE \cite{Pascual2019LearningTasks} go beyond defining a single self-supervision pretext task to combines several losses, each predicting or classifying a different property of the input.


\subsubsection{Datasets}
While it is not as strong as in the text modality, there is still a trend for newer models to train on larger and larger datasets. Small datasets historically used for model model training, are now reserved for downstream evaluation, with contemporary mehods being pre-trained on 
LibriSpeech \cite{Panayotov2015Librispeech:Books} containing 960 hours of speech from audiobook readings and Libri-light \cite{Kahn2019Libri-Light:Supervision}, a much larger dataset (60K hours) of similar audiobook recordings.

\cut{
While the focus so far is mainly on modelling English language speech, there has been training and evaluation done on large datasets in Mandarin \cite{Jiang2019ImprovingPre-training}, namely HKUST/MTS \cite{Liu2006HKUST/MTS:Corpus} and AISHELL-1 \cite{Bu2017AISHELL-1:Baseline}.
}

\cut{
\minikeypoint{Target}
The main task of interest to the self-supervised community in this modality has been speech recognition on LibriSpeech \cite{Panayotov2015Librispeech:Books}, WSJ \cite{Paul1992TheCorpus}, TIMIT \cite{Garofolo1993TIMITCorpus} or in noisier environments in DIRHA \cite{Ravanelli2015TheEnvironments}. The majority of papers do not consider a wider range of datasets for transfer. As performance further improves, the need for wider evaluation might emerge as it has in other modalities \cite{Ericsson2021HowTransfer}. For now only a few methods has assessed such broader transfer tasks, e.g.~speaker identification on VCTK \cite{Veaux2017CSTRToolkit} and emotion classification on INTERFACE \cite{Hozjan2002InterfaceDatabase}.
}

\subsubsection{Applications}
A notable success in Speech was shown by Wav2Vec 2.0 \cite{Baevski2020Wav2vecRepresentations} which used transformers+masked prediction SSRL on 53k hours of unlabelled data prior to fine-tuning a downstream speech recognition system. This was subsequently able to surpass prior state of the art ASR performance with 10-fold less supervised data than used before, and approach state of the art with 100-fold less supervised data than before. Albeit at the cost of 660 GPU-days of SSRL compute, this is a dramatic improvement in data efficiency.

In terms of more general time-series data representation learning, masked prediction methods based on transformer architecture have been shown to match supervised state of the art in a variety of diverse benchmarks a suite of benchmarks in diverse application areas \cite{franceschi2019unsupervised}.  

{A major application area for self-supervised time-series analysis is medical data, where annotations are hard to collect. There has been progress in applying SSRL to EEG and ECG data 
\cite{Sarkar2020Self-supervisedRecognition,Cheng2020Subject-AwareBiosignals}. For example using transformation prediction SSRL prior to training ECG-based emotion recognition \cite{Sarkar2020Self-supervisedRecognition} and contrastive instance discrimination SSRL prior to learning downstream EEG-based motor movement classification and ECG-based anomaly detection \cite{Cheng2020Subject-AwareBiosignals}.} 
In terms of time-series forecasting, transformers based on sequential masked prediction pretext have significantly outperformed conventional autoregressive models in predicting disease transmission \cite{wu2020deepForecast}.

\cut{
As per the text domain (Section~\ref{ssec:text:apps}), speech analysis suffers from a major bottleneck in terms of lack of annotated data, or even un-annotated data, from the majority of low-resource languages. Recent work building on CPC has shown that SSRL pre-training can learn powerful general purpose representations that can be fine-tuned to comparatively data-efficiently to achieve automatic speech recognition in low-resource languages \cite{Riviere2020UnsupervisedLanguages}.
}


\subsection{Graphs} 
Graph structured data is ubiquitous in the networked world, and supports a diverse array of tasks including node, edge, and graph classification. These tasks should all be informed by both node/edge features where available, and graph connectivity. Graph Neural Networks \cite{Kipf2017Semi-SupervisedNetworks} have advanced all these tasks significantly, especially where massive labelled data is available. Thus a large body of work on self-supervised graph representation learning has emerged to facilitate downstream GNN-based tasks. 

Graph-based SSRL can be somewhat unique in several aspects. Depending on whether the ultimate task of interest requires node-level or graph-level predictions, methods may focus on learning node-level \cite{Velickovic2019DeepInfomax,Hamilton2017InductiveGraphs,Hu2020GPT-GNN:Networks} or graph-level \cite{Sun2020InfoGraph:Maximization} representations, or both \cite{Hu2020StrategiesNetworks}. Graph-based methods also differ in whether they are oriented at training on a \emph{set} of graphs \cite{Velickovic2019DeepInfomax,Sun2020InfoGraph:Maximization} (cf: set of images or audio clips in other modalities), or on a single large graph \cite{Grover2016Node2vec:Networks}.

\subsubsection{Methods}
Early shallow methods for self-supervised graph representation learning used NLP-inspired masked prediction approaches to learn node-embeddings, for example based on random walks on the graph  \cite{Grover2016Node2vec:Networks}. Much as shallow word embeddings have been eclipsed by deep language models in NLP, newer graph-representation learning architectures that focus on graph {convolutional networks or self-attention} have driven progress in this modality. 

In terms of self-supervision objectives, most work in this area falls into masked prediction and instance discrimination categories. Several recent methods optimise mutual information-based instance discrimination objectives, with  DGI \cite{Velickovic2019DeepInfomax} and InfoGraph \cite{Sun2020InfoGraph:Maximization} performing contrastive instance discrimination between pairs of nodes/patches and whole graphs. 
Masked prediction pretexts were used both by classic shallow methods \cite{Grover2016Node2vec:Networks}, as well as recent deep approaches such as GPT-GNN \cite{Hu2020GPT-GNN:Networks}. A minority of approaches  have applied transformation-prediction methods -- such as GraphTER \cite{Gao2020GraphTER:Transformations}, where node-wise transformations are applied and and predicted by a GNN.



An important dichotomy in graph-based representation learning is between transductive and inductive graph representation learning methods. The majority of methods are transductive, in that they learn embeddings specifically for nodes seen during, and so are primarily relevant in applications where the downstream task uses the same graph data as is used for pre-training. This is analogous to how the word2vec algorithm \cite{Mikolov2013DistributedCompositionality} in language learns embeddings for words in its training set, but cannot produce embeddings for unseen words. A minority of methods are inductive \cite{Hamilton2017InductiveGraphs,Hu2020GPT-GNN:Networks} in that they learn embedding functions that do not depend on a specific choice of input graph, and thus can be transferred to new target nodes or graphs.

\subsubsection{Datasets}
Since the graph structured data occurs so pervasively, it covers a wide range of data types tasks. Major examples include social  \cite{Hamilton2017InductiveGraphs}, citation  \cite{Zhang2019OAG:Graphs}, chemical \cite{Wu2018MoleculeNet:Learning} and biological networks \cite{Zitnik2017PredictingNetworks}. Because there are many different kinds of graphs with different structures and sizes there is no one-size-fits-all source dataset which consistently improves transfer as in many of the previously discussed modalities. It instead depends on the tasks of interest.

For learning in the transductive setting, the pre-training must necessarily be done on the same graph as the testing, thus limiting the transfer task-transfer and not domain-transfer. For the inductive setting, the source data can differ from target but in most evaluation cases the test set consists of nodes that were hidden from the training graph \cite{Hu2020GPT-GNN:Networks} or unseen graphs from the same underlying dataset \cite{Zitnik2017PredictingNetworks}. Like in other modalities we have seen increasingly large graphs being used for pre-training, like the Amazon Review Recommendation data \cite{Ni2019JustifyingAspects} with 113 million nodes or Open Academic Graph (OAG) \cite{Zhang2019OAG:Graphs} consists of over 178 million nodes and 2 billion edges. 

\cut{
\minikeypoint{Target}
The two most commonly used target tasks are node classification -- where each node of a single graph is assigned into one of several classes -- and graph classification -- where a each of several graphs is assigned a class. Common datasets for node classification include CiteSeer \cite{Sen2008CollectiveData} and PubMed \cite{Sen2008CollectiveData} in the transductive setting, and Reddit \cite{Hamilton2017InductiveGraphs} and PPI \cite{Zitnik2017PredictingNetworks} in the inductive setting. For graph classification there is MUTAG \cite{Debnath1991Structure-ActivityHydrophobicity}, IMDB-Binary/Multi \cite{Yanardag2015DeepKernels}, MoleculeNet \cite{Wu2018MoleculeNet:Learning} and Reddit \cite{Hamilton2017InductiveGraphs}.
}

\subsubsection{Applications}

Self-supervised graph-based representation learning is expected benefit all graph-based prediction applications where data is limited. This is especially the case in computational chemistry  and biology applications, where graphs and associated annotations may correspond to molecules and corresponding molecular properties. In such applications data are intrinsically hard to collect, but predicting graph properties can significantly impact tasks such as drug discovery and material discovery \cite{Wu2018MoleculeNet:Learning,Zitnik2017PredictingNetworks}. In computer vision, using LIDAR rather than RGB sensors leads to observations represented as point clouds or graphs rather than conventional images. In this case self-supervised graph representation learners such as GraphTER \cite{Gao2020GraphTER:Transformations} have led to excellent performance in object segmentation (i.e., node classification) and classification (i.e., graph classification).

\section{Discussion}\label{sec:discuss}
\subsection{Pre-training Cost} The pre-training cost of different SSRL methods is not consistently documented, and hardware platform/GPU differences make them hard to compare quantitatively. Nevertheless, it is clear that we can see that state of the art methods in computer vision, speech and text  (Tables~\ref{tab:methods} and \ref{tab:datasets}) require massive resources on the order of 100s of GPU-days for training on ImageNet, LibriSpeech, and Wikipedia corpora respectively. The general purpose pre-trained nature of these representations may amortise this cost somewhat, by enabling many downstream problems to be solved with the same representation. This has largely been the case in the text modality  where there has been strong success fine-tuning generic pre-trained models to diverse tasks \cite{Devlin2019BERT:Understanding}. However this may not be possible in other modalities such as graphs which may require transductive training, or vision where domain-specific pre-training may be necessary for data very different to ImageNet such as hyperspectral imagery or volumetric MRI. In this case pre-training cost poses an accessibility barrier to modestly resourced organisations, and an environmental issue \cite{schwartz2019greenAI} due to its energy requirement.  While there is also tremendous research activity in developing more efficient pre-training algorithms, the net cost of pre-training is trending upward due to the fact that bigger datasets and bigger network architectures have systematically led to better performance. 


\subsection{Data Requirement and Curation}
For text \cite{Brown2020LanguageLearners} and speech \cite{Baevski2020Wav2vecRepresentations} the literature unambiguously shows that thus far performance increases consistently with ever larger datasets. In the case of text, this result further seems to be relatively insensitive to the degree of curation of the data.

For images, the majority of recent work still uses  still uses ImageNet with its 1.28 million images as the source set \cite{Caron2020UnsupervisedAssignments,Grill2020BootstrapLearning}. However, a number of studies have shown that using larger pre-training datasets \cite{Thomee2015YFCC100M:Research,Mahajan2018ExploringPretraining}  benefits to transfer performance \cite{Mahajan2018ExploringPretraining,Goyal2019ScalingLearning}, with feature quality growing logarithmically with data volume \cite{Goyal2019ScalingLearning}. For video pre-training, the state-of-the-art models use the increasingly large YouTube8M-2 \cite{YouTube-8M:Research} and HowTo100M  \cite{Miech2019HowTo100M:Clips}
with combined video play-times of 13 and 15 years respectively.

The vision of SSRL is to enable representation learning on easily obtained uncurated data. However, for benchmarking purposes (especially in vision and audio and graphs, but less so in text), methods are often actually trained on curated data while ignoring the labels. It is not clear how much existing algorithm design is overfitted to these curated datasets, and if the relative performance of different methods is maintained when real uncurated data is used instead. For example in computer vision, most pre-training is performed on ImageNet, which is large and diverse, yet uniformly focused on individual objects. If this was replaced with scene images with multiple cluttered objects, then typical instance discrimination tasks like mapping two different crops of one image to the same identity could create false positive pretext label noise that maps different semantic objects to the same representation \cite{Purushwalkam2020DemystifyingBiases}. We are beginning to see new SSRL methods designed for data with different statistics such as cluttered images \cite{Purushwalkam2020DemystifyingBiases}.

\subsection{Architecture Choice and Deployment Costs}\label{ssec:discussArchitecture}
For both image \cite{Chen2020BigLearners,Goyal2019ScalingLearning} and text \cite{Devlin2019BERT:Understanding} analysis, the trend has been that bigger architectures lead to better representation performance, especially when coupled with extremely large pre-training datasets, and challenging pretext tasks \cite{Goyal2019ScalingLearning}. This is welcome from the perspective of near `automatic' performance improvement as datasets and compute capabilities grow. However, it does pose a concern for deployment of the resulting models on resource-constrained or embedded devices with limited memory and/or compute capability, which may limit the benefit of this line of improvement for such applications.

A standard approach to alleviate this issue is to perform SSRL of large models as usual followed by using unlabelled data to perform post-training \emph{distillation} of the large self-supervised model into a smaller more compact but similarly performant student model. For example, in vision this has been demonstrated to compress a ResNet-152$\times$3 model to a ResNet-50 of similar performance \cite{Chen2020BigLearners}; in text a 109M parameter/22.5GFLOP BERT model can be distilled to a 14.5M/1.2GFLOP BERT model with similar performance \cite{Jiao2020TinyBERT:Understanding}.

\subsection{Transferability}
The vision of SSRL is to produce features that transfer to a wide range of downstream tasks. The extent to which this has been realised varies by discipline/modality. In vision this is on its way with many studies evaluating transfer performance \cite{Ericsson2021HowTransfer,Goyal2019ScalingLearning}, 
but no single benchmark has yet been widely agreed. Recognition has been the most common scene of transfer assessment but recently detection and dense prediction have also been embraced \cite{Caron2020UnsupervisedAssignments,Grill2020BootstrapLearning,Ericsson2021HowTransfer}. However, ImageNet Top-1 accuracy is still the main metric used in model comparisons. As reported by \cite{Ericsson2021HowTransfer}, this metric shows high correlation with downstream recognition performance. Their results for detection and dense prediction, however, show markedly lower correlations, indicating that current SSRL methods are not optimised for such a broad transfer \cite{Ericsson2021HowTransfer}. For practitioners with new data and tasks, this means that the best performing SSRL model on ImageNet can safely be adapted to recognition tasks. However, if the task differs then more models need to be considered. Additionally, if the images of the target domain are unstructured or exhibit different properties to ImageNet images, then further caution must be taken when choosing a pre-trained model. This is further expanded on by \cite{MacAodhaBenchmarkingCollections} who show how SSRL models fail to compete with supervision on in-the-wild datasets containing plant and animal species, contrasting what has been found for curated datasets \cite{Ericsson2021HowTransfer}.

In video and multi-modal settings, common transfer evaluation considers transfer from large source datasets such as Kinetics \cite{Kay2017TheDataset} to standard target datasets such as UCF101. State of the art methods  successfully leverage large source datasets and approach but do not yet outperform supervised pre-training \cite{Han2020Self-supervisedLearning}. 
Nonetheless, there has been an uptake of SSRL methods in applications such as 
tracking \cite{wang2019cycleTime} and detection. 

In text, the field has matured more already. Here, several broad benchmarks such as SuperGLUE \cite{Wang2019SuperGLUE:Systems}  
are regularly used to monitor progress. The main mode of transfer in NLP has long been to fit a linear model or fine-tune a SSRL model like BERT \cite{Devlin2019BERT:Understanding}, and on many tasks on the above benchmarks fine-tuned SSRL models achieve top results.
The recent GPT-3 \cite{Brown2020LanguageLearners} has shown that huge SSRL models can achieve competitive performance via few-shot adaptation instead of full fine-tuning, especially on language modeling and question answering. In summary, text models exhibit relatively high transferability with SSRL pre-training dominating in a broad range of downstream tasks.

In speech and time-series the focus has so far been narrow, with only a few tasks and datasets forming the evaluation landscape. These cover phoneme recognition and occasionally speaker identification or emotion classification. Most work focuses on the English language speech both for pre-training and transfer. However very rapid progress is currently being made in multi-lingual \cite{Kannan2019Large-ScaleModel} speech models and cross-lingual transfer \cite{Riviere2020UnsupervisedLanguages}, so prospects for transferability seem promising.

{The current state of the graph modality is that transferability is good to unseen nodes within the same graph and to unseen graphs within the same dataset, e.g.~ PPI \cite{Zitnik2017PredictingNetworks}. However, there is little information to suggest transfer across graph types, like chemical-to-biological or citation-to-social, currently has any benefit.}

\subsection{Choosing the Right Pretext Task}
As we have seen, the four families of pretext tasks can be applied to all the different modalities. But because self-supervised pretexts rely on exploiting the structure of data, which in turn differs significantly across modalities, their efficacy can vary substantially across modalities. One clear such trend is that masked prediction is ubiquitous in the text modality \cite{Mikolov2013DistributedCompositionality,Devlin2019BERT:Understanding,Brown2020LanguageLearners}, with other tasks being significantly less effective. And when other tasks are used, they are often complementary to a masked prediction loss \cite{Lewis2020BART:Comprehension}. In images, masked prediction and transformation prediction have been tried in various forms and drove initial progress, but the most recent advances in these modalities have been driven by instance discrimination \cite{Chen2020BigLearners,Han2020Self-supervisedLearning} and clustering \cite{Caron2020UnsupervisedAssignments,Alwassel2020Self-SupervisedClustering}. 
However, transformation prediction is still seeing success in videos, presumably because of of the rich spatio-temporal information to be exploited. Finally, while there may be a dominant pretext strategy for a given modality, it is common that a suitably designed combinations of pretexts applied in a multi-task manner can improve performance compared to a single pretext \cite{Lewis2020BART:Comprehension}.

Picking a pretext based on the bulk of successes for the modality of interest is a good start. However to further inform choice, one can further consider the assumptions which underlie each family of methods. Masked prediction relies on context being enough to fill in missing parts of a datapoint. Transformation prediction relies on each datapoint possessing a canonical view. Instance discrimination relies on each datapoint representing a unique semantic example, distinguished from all other datapoints in the training set, which may not hold for cluttered images as discussed above. It is notable that clustering requires no strong assumptions other than the existence of meaningful similarities by which to group the data into a certain number of clusters. Therefore, if little is known of the structure of the data, then a method based on clustering may be a good start.

A final consideration when selecting a pretext task is \emph{what properties do we want in our representations?} If our data modality is images and we are interested in exploiting the orientation of objects in our data, do we want our representations to vary with orientation -- in which case we might want to use a transformation prediction method like \cite{Gidaris2018UnsupervisedRotations} -- or do we want all orientations of the input to produce the same output -- in which case we might instead choose an instance discrimination method that uses rotation-based augmentation. This question of \emph{equivariance} or \emph{invariance} can greatly impact the downstream performance of certain tasks. For example, a visual object classification task might benefit from invariance to spatial translation but a detection task would need this information to be preserved in order to correctly predict object locations.

If there are no specific downstream task in mind a-priori and therefore no known required properties that must be learned, the ideal selection is not clear. In this case we want to use the method which best captures the core information in our data which has the best chance of being of use for later tasks. Finding such pretext tasks can be considered the main aim of the self-supervised representation learning field of research.

\cut{
Given the variety of data seen in the graph modality, the choice of method may be dependent both on the type of graph we wish to learn on and the type of transfer we will consider. For graphs with rich attribute annotations, a masked attribute prediction method might work well. For graphs lacking attribute information, there are methods designed for this setting too \cite{Qiu2020GCC:Pre-Training}. If the target task contains nodes not seen during training, then an inductive setting must be used, and an appropriate method chosen.
}

\subsection{Self-supervised vs Semi-supervised}
In cases where the source and target datasets are the same or similar in content and label-space, then both \emph{semi-supervised} and \emph{self-supervised} approaches can potentially apply (Section~\ref{sec:background}). Since both families of methods are making rapid progress and there have been few direct comparisons, it is not yet clear if/when one family should be preferred. However, since SSRL deals with initialisation and SSL deals with refinement, the two strategies can in principle both be applied to one learning problem. There has yet been very little investigation into the extent to which these strategies can be complementary and further boost performance when used together. A preliminary result in computer vision suggests not \cite{zoph2020rethinking}. However, preliminary results in 
text \cite{du2020selftrainingSSRL} suggest that SSL and SSRL can be synergistic when used together.

\subsection{Other Benefits of SSRL}
While we have mainly focused on the benefits of SSRL with respect to accuracy in the low and few-shot data regime, there are several other potential benefits: (i) The computational cost of fine-tuning a self-supervised model tends to be lower than training from scratch (though comparable to fine-tuning a supervised pre-trained feature). (ii) If the supervised target task suffers from label-noise, training leads to much worse performance compared to using clean labels. However SSRL also increases resilience to such label noise \cite{hendrycks2019ssl},  which often occurs in practice. (iii) Given a trained system, SSRL can also improve robustness of image recognition to adversarial attack, as well as common corruptions such as blur, noise, and compression artefacts \cite{hendrycks2019ssl}. (iv) Furthermore SSRL leads to better calibrated probabilities \cite{Ericsson2021HowTransfer,hendrycks2019ssl}, which can be used to drive abstention of automated predictions, or out-of-distribution detection \cite{hendrycks2019ssl}. (v) Finally, in terms of model interpretability feature extractors trained by self-supervision tend to lead to more reasonable and interpretable attention maps \cite{Ericsson2021HowTransfer}.

\updated{
\subsection{Recommendations for Future Work}
\begin{itemize}
    \item Develop wider benchmarks. Several of the modalities we look at have a few standard downstream tasks that are consistently evaluated against. This creates a bias towards making new methods that optimise only for those particular tasks. Instead we should create benchmark suites that study the performance of pre-trained models across a wide range of tasks within a modality. This has been successfully done in NLP and has driven progress and made sure it benefits many areas of the field \cite{Wang2019SuperGLUE:Systems}, but such standardised benchmarks are lacking in the other modalities we have considered.
    \item Do not only focus on tracking task performances in these benchmarks but also track other feature properties like social biases to obtain broader understanding of how these models behave. Progress on reducing such biases can only really be done if we know about and can quantify them.
    \item Be wary of relying only on scale to improve performance. As we use larger and larger datasets to train these models, we know less and less about the data itself as there is very little human oversight in the data collection process. By developing methods that are more data efficient -- i.e.~don't need billions of instances to learn -- we can create models that are easier to understand and control. Additionally, as we develop larger models their carbon footprint grows significantly \cite{schwartz2019greenAI}. Make sure that the efficiency of training these models is tracked in common benchmarks.
    \item Do not get stuck on training on only one specific source dataset as this will bias the type of methods that are created. As an example, the highly curated and single-centred-object-style of ImageNet has led to a particular style of data augmentation and instance discrimination. However, it has been shown that on less curated `in-the-wild' images, these methods underperform. By continuously considering different types of source datasets we get a better picture of when and where a method works.
\end{itemize}
}

\section*{Acknowledgements}
This research was partially supported by the Engineering and Physical Sciences Research Council (EPSRC) Grant number EP/S000631/1 and the MOD University Defence Research Collaboration (UDRC) in Signal Processing; EPSRC Centre for Doctoral Training in Data Science, funded by
EPSRC (grant EP/L016427/1) and the University
of Edinburgh; and EPSRC grant EP/R026173/1.

\bibliographystyle{IEEEtran}
\bibliography{references}

\end{document}